\pdfoutput=1
\documentclass[journal]{IEEEtran}
\usepackage[colorlinks=true,linkcolor=blue,citecolor=blue,urlcolor=blue]{hyperref}

\usepackage[cmex10]{amsmath}
\usepackage{amssymb}
\usepackage{graphicx}
\usepackage[table]{xcolor}
\usepackage{tabularray}
\definecolor{gray9}{gray}{0.9}
\usepackage[style=ieee,backend=biber,sorting=none]{biblatex}
\UseTblrLibrary{booktabs}
\DeclareFieldFormat{bibliography}{\setlength{\baselineskip}{1.0\baselineskip}#1}

\begin{document}
\title{SPMR: Bridging Set Prediction and Multi-Positive Recognition for Fine-Grained Object Detection in Remote Sensing Images}

\author{Wentao~Li,
  Danpei~Zhao,~\IEEEmembership{Senior Member,~IEEE,}
  Bo~Yuan,
  Tian~Li,
  and~Zhenwei~Shi,~\IEEEmembership{Senior Member,~IEEE}%
  \thanks{This work was supported by the National Natural Science Foundation of China under Grant 62271018. (Corresponding author: Danpei Zhao.)}%
  \thanks{Wentao Li, Danpei Zhao, Bo Yuan, Tian Li, and Zhenwei Shi are with the Department of Aerospace Intelligent Science and Technology, School of Astronautics, Beihang University, Beijing 102206, China, and the Key Laboratory of Spacecraft Design Optimization and Dynamic Simulation Technology, Ministry of Education, Beijing 102206, China. Danpei Zhao, Bo Yuan and Tian Li are also with the Tianmushan Laboratory, Beihang University, Hangzhou 311115, China. (e-mail: canoe@buaa.edu.cn; zhaodanpei@buaa.edu.cn; yuanbobuaa@buaa.edu.cn; lit@buaa.edu.cn; shizhenwei@buaa.edu.cn).}}

\markboth{IEEE TRANSACTIONS ON GEOSCIENCE AND REMOTE SENSING}%
{Li \MakeLowercase{\textit{et al.}}: SPMR: Bridging Set Prediction and Multi-Positive Recognition for Fine-Grained Object Detection in Remote Sensing Images}

\makeatletter
\def\ps@IEEEtitlepagestyle{%
  \def\@oddhead{\hbox{}\scriptsize\leftmark \hfil \thepage}%
  \def\@evenhead{\scriptsize\thepage \hfil \leftmark\hbox{}}%
  \def\@oddfoot{\raisebox{10pt}[0pt][0pt]{\parbox[t]{\textwidth}{\centering\scriptsize\setlength{\baselineskip}{7.5pt}\color{black!75}%
    \textcopyright~2026 IEEE. Personal use of this material is permitted. Permission from IEEE must be obtained for all other uses, in any current or future media, including reprinting/republishing this material for advertising or promotional purposes, creating new collective works, for resale or redistribution to servers or lists, or reuse of any copyrighted component of this work in other works.}}}%
  \def\@evenfoot{\@oddfoot}%
}
\makeatother

\maketitle
\begin{abstract}
  Fine-grained object detection (FGOD) in remote sensing requires accurate localization and fine-grained recognition under subtle visual differences. Current query-based detectors enable NMS-free inference through one-to-one set prediction, but their sparse positive supervision makes fine-grained recognition difficult. To mitigate this limitation, we propose SPMR, a framework that combines set prediction with multi-positive recognition for remote sensing FGOD. SPMR separates class-agnostic proposal generation from fine-grained recognition and applies different supervision to the two stages. The Sparse Oriented Proposal Network (SOPN) generates sparse class-agnostic oriented proposals and estimates their objectness under one-to-one set supervision. The Query-Aware Refinement Head (QARH) then performs proposal-level fine-grained recognition with quality-aware multi-positive supervision. During inference, Dual-Stage Consensus Scoring (DSCS) fuses the objectness score from SOPN with the fine-grained class score from QARH to produce final predictions without NMS. Experiments on MAR20, ShipRSImageNet, and FAIR1M show that SPMR outperforms representative NMS-free oriented detectors and remains competitive with strong two-stage FGOD methods. Specifically, SPMR obtains 85.82 AP on MAR20, 75.40 AP on ShipRSImageNet, and 52.12 AP on FAIR1M-v2.0. Code is available at: \url{https://github.com/AIRVIC-Lab/SPMR}.
\end{abstract}

\IEEEpeerreviewmaketitle

\section{Introduction}
\IEEEPARstart{F}{ine-grained} object detection (FGOD) in remote sensing imagery aims to localize objects and recognize fine-grained classes, such as distinguishing an \textit{Airbus A350} from a \textit{Boeing 747}. With the increasing availability of high-resolution satellite and aerial images, FGOD has been used in many applications, including maritime surveillance \cite{shiprs,osfd,dosr,sis}, aviation management~\cite{acd,mar20,iscl,gicnet}, urban planning~\cite{ubc}, and disaster response~\cite{fdd}. Compared with general object detection, it is mainly constrained by two coupled challenges: semantic confusion and task interference.

\begin{figure}[!t]
  \begin{center}
    \includegraphics[width=0.48\textwidth]{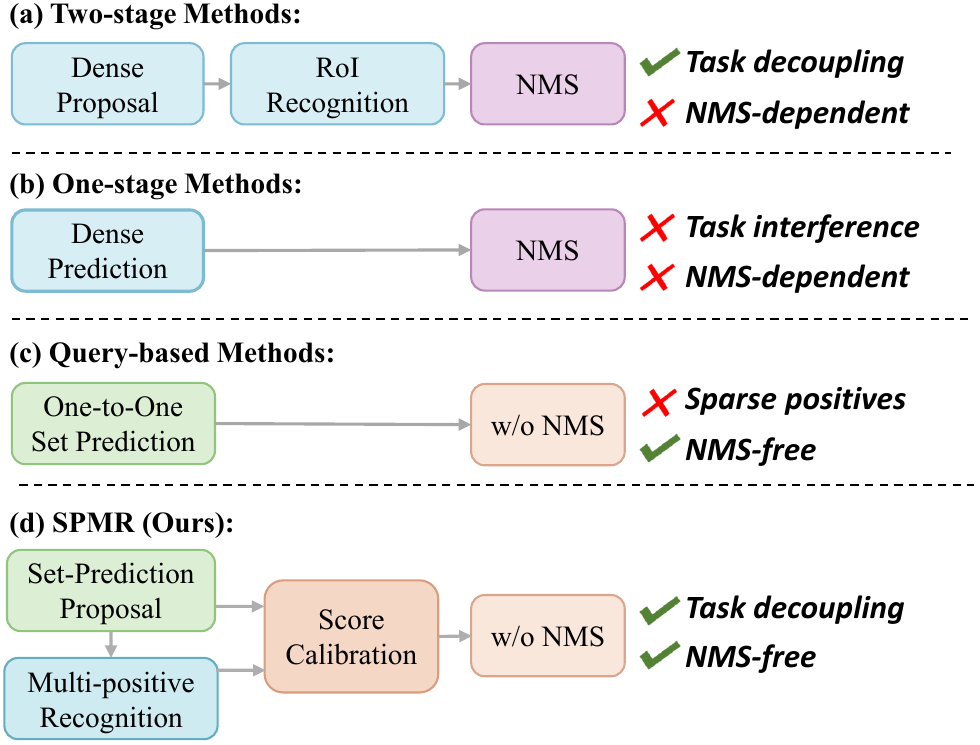}
  \end{center}
  \caption{Comparison of detection paradigms for FGOD.
    (a) Two-stage detectors use dense proposals and RoI-level recognition followed by NMS.
    (b) One-stage detectors perform dense joint prediction followed by NMS.
    (c) Query-based detectors use sparse one-to-one queries for NMS-free prediction.
    (d) SPMR generates sparse class-agnostic proposals under one-to-one supervision and performs proposal-level multi-positive recognition with score fusion for NMS-free inference.}
  \label{fig:intro}
\end{figure}

A common challenge in FGOD is \textit{semantic confusion}: objects from different fine-grained classes with subtle visual differences may appear highly similar in top-down remote sensing images, while instances of the same class can vary substantially in scale, orientation, and background. This large intra-class variation and small inter-class difference make discriminative recognition difficult.

Beyond semantic confusion, FGOD also suffers from \textit{task interference}, because foreground-background classification, box regression, fine-grained recognition, and duplicate suppression impose different optimization preferences within a unified detection pipeline. Foreground classification and duplicate suppression favor a small number of confident object hypotheses, box regression emphasizes geometric accuracy, whereas fine-grained recognition depends on subtle local appearance differences. As a result, a proposal that is geometrically accurate or highly confident as foreground may not always provide the most discriminative local features for fine-grained recognition.

Most recent FGOD methods \cite{pcldet, sfrnet, petdet, drnet} are built on two-stage detectors. As summarized in \autoref{fig:intro}, existing detection paradigms differ in how they distribute localization, fine-grained recognition, and duplicate suppression. Conventional two-stage detectors separate dense proposal generation from RoI-level recognition, which alleviates part of the interference between foreground-background classification and fine-grained recognition, but they still rely on NMS to remove duplicated predictions. By contrast, one-stage detectors perform dense prediction directly on feature maps, where foreground classification, box regression, fine-grained recognition, and duplicate suppression are more tightly coupled, and NMS is also required in post-processing.

Query-based detectors, represented by DETR~\cite{detr}, provide an alternative paradigm. They formulate object detection as set prediction and use one-to-one matching to avoid NMS during inference. However, as illustrated in \autoref{fig:intro}(c), the same sparse query set is expected to localize objects, suppress duplicates, and recognize fine-grained classes. This design removes pairwise suppression at inference, but it also provides limited positive supervision for learning subtle inter-class differences. When such detectors are directly used for remote sensing FGOD, their accuracy often remains below that of strong two-stage detectors.

We consider this limited positive supervision a structural limitation of directly applying one-to-one set prediction to FGOD. Under one-to-one matching, each ground-truth object is matched to a single object query, and only the matched query receives positive fine-grained classification supervision for that object. Other queries may also produce localized predictions for the same object, but these predictions are not used as additional positives. Queries left unmatched by the assignment are instead optimized as background for classification. Consequently, potentially informative localized predictions cannot provide additional positive supervision for learning class-discriminative features, making it more difficult to distinguish visually similar fine-grained categories. This supervision mismatch motivates retaining one-to-one matching for class-agnostic objectness estimation and localization while using overlapping proposals as multiple positives for proposal-level fine-grained recognition.

To this end, we propose SPMR, a set-prediction and multi-positive recognition framework for remote sensing FGOD. SPMR follows the proposal-to-recognition flow of a conventional two-stage detector. However, it generates proposals through set prediction and does not rely on NMS to remove duplicate predictions. As shown in \autoref{fig:intro}(d), the Sparse Oriented Proposal Network (SOPN) uses a fixed number of object queries to generate class-agnostic oriented proposals under global one-to-one matching. Unlike a dense RPN, SOPN passes the complete proposal set to QARH without proposal-level NMS. The Query-Aware Refinement Head (QARH) then performs multi-positive fine-grained recognition and box refinement on these proposals. Dual-Path Auxiliary Supervision (DPAS) provides additional localized proposals during training without changing the inference path. Dual-Stage Consensus Scoring (DSCS) combines SOPN objectness and QARH class scores for final ranking without NMS.

Experiments on MAR20, ShipRSImageNet, and FAIR1M show that SPMR outperforms representative NMS-free oriented detectors and remains competitive with well-established two-stage FGOD methods. On MAR20 and ShipRSImageNet, it improves over DINO-O by 10.46 AP and 5.08 AP, respectively. On FAIR1M-v2.0, it improves over the recent NMS-free method QEDetr~\cite{qedetr} by 3.53 AP under the same multi-scale setting.

The main contributions of this work are summarized as follows:
\begin{enumerate}
  \item We propose SPMR, a role-separated FGOD framework that combines one-to-one set prediction for class-agnostic proposal generation with multi-positive supervision for fine-grained recognition, enabling NMS-free inference with richer recognition supervision.
  \item We propose SOPN and QARH to instantiate this role separation. SOPN generates a sparse proposal set, whereas QARH integrates query and RoI features for fine-grained recognition. DSCS combines their complementary scores for NMS-free ranking.
  \item Extensive experiments on MAR20, ShipRSImageNet, and FAIR1M demonstrate that SPMR outperforms representative query-based oriented detectors and remains competitive with strong two-stage FGOD methods.
\end{enumerate}

\section{Related Work} \label{sec:related}
\subsection{Query-based Methods for Object Detection}
Query-based object detection has been extensively studied since DETR~\cite{detr} formulated detection as a set prediction problem and removed the need for NMS. Existing approaches can be roughly grouped into query design and training supervision. The first line improves the spatial prior and feature interaction of object queries. Deformable-DETR~\cite{deform} uses multi-scale deformable attention to speed up convergence. Conditional-DETR~\cite{conditional}, DAB-DETR~\cite{dab}, and Anchor-DETR~\cite{anchor} introduce anchor-like spatial priors into object queries. Some methods further refine queries with RoI features~\cite{sam,hpr}, model object relations~\cite{relation}, or simplify decoder components for better efficiency~\cite{dac,msdetr,mrdetr}.

Optimization-oriented methods focus on richer supervision and more stable training. DN-DETR~\cite{dn} and DINO~\cite{dino} introduce denoising queries to provide direct reconstruction targets. H-DETR~\cite{hdetr} and Group-DETR~\cite{group} adopt one-to-many assignment during training, usually by duplicating ground truths or adding auxiliary query groups. Align-DETR~\cite{align} and Stable-DINO~\cite{stable} further study the mismatch between classification and localization from the perspective of task alignment. These methods show that sparse set prediction can benefit from additional supervision beyond the final one-to-one assignment.

Despite these advances, auxiliary assignments in DETR variants do not change the role of the final prediction queries: in the main prediction branch, each object is still supervised by a single matched query. For FGOD, this leaves the same query responsible for competing tasks, including foreground discrimination, duplicate suppression, localization, and fine-grained recognition. SPMR therefore keeps one-to-one set prediction for class-agnostic proposal generation and trains the proposal-level recognition path with multi-positive supervision. The same recognition path is retained at inference.

\subsection{Oriented Object Detection}
Oriented object detection is critical for remote sensing because oriented bounding boxes reduce background interference for objects with arbitrary directions. Early CNN-based detectors can be roughly divided into two-stage and one-stage methods. Two-stage methods~\cite{r2cnn,rrpn,scrdet,roitrans,aopg,gv,redet,orcnn} generate proposals first and then refine them, but they still depend on NMS. One-stage methods predict dense boxes directly. Many of them improve feature alignment~\cite{r3det,s2anet,ridet,cfc,tioe} or design better angle representations and regression losses~\cite{gwd,kld,psc,acm,gaucho}.

Recent studies increasingly extend DETR-like detectors to oriented detection~\cite{ao2,d2q,oformer,arsdetr,odetr,rhino,ro2}. These methods address oriented detection from several perspectives. Some works improve geometric query modeling, such as AO2-DETR~\cite{ao2} with rotation-invariant proposal refinement, ARS-DETR~\cite{arsdetr} with rotational deformable attention, OrientedFormer~\cite{oformer} with oriented cross-attention, and RO2-DETR~\cite{ro2} with rotation-equivariant modeling. Other works address angle periodicity, boundary discontinuity, or localization quality through classification-based angle prediction~\cite{arsdetr}, Hausdorff distance~\cite{rhino}, Gaussian-based encoding~\cite{oformer}, point-axis representation~\cite{odetr}, or query-quality enhancement~\cite{d2q}.

While these methods improve oriented localization and angle modeling, their main focus remains geometric prediction. In FGOD, however, visually similar objects also require proposal-level features that preserve local discriminative evidence. SPMR addresses this complementary requirement by using SOPN for class-agnostic oriented proposal generation and QARH for fine-grained recognition from multiple proposal samples.

\subsection{FGOD in Remote Sensing Images}
FGOD in remote sensing images has attracted increasing attention because aircraft, ships, and other man-made objects often exhibit subtle inter-class differences~\cite{acd,osfd,iscl,fgfe,gicnet,sis,pcldet,sfrnet,petdet,drnet}. Prior studies can be organized by how they address semantic confusion. Feature-enhancement methods improve discriminative representations for fine-grained classes. ACD~\cite{acd}, OSFD~\cite{osfd}, ISCL~\cite{iscl}, and GICNet~\cite{gicnet} improve class-sensitive representations for aircraft or ship recognition. SFRNet~\cite{sfrnet} refines spatial and channel features separately, and DRNet~\cite{drnet} introduces a parallel branch with a confusion minimization loss.

Metric and contrastive learning methods further enlarge inter-class separability. PCLDet~\cite{pcldet} uses prototypical contrastive learning to improve class discrimination. Such designs are effective in improving fine-grained representation, and most of them are built on two-stage oriented detectors with dense proposals and NMS. A related direction studies the conflict between localization and recognition. PETDet~\cite{petdet} discusses FGOD from a multi-task learning perspective, but it still follows a conventional two-stage CNN pipeline.

Query-based detectors provide a possible way to remove NMS from FGOD. QEDetr~\cite{qedetr}, built on DINO~\cite{dino} and PSC~\cite{psc}, improves query discriminability with IoU-aware supervision, showing the potential of NMS-free FGOD. However, it still follows the common strategy of strengthening the same query prediction head for both localization and fine-grained classification. BHCL~\cite{bhcl} introduces balanced hierarchical contrastive learning for query-based FGOD and decouples classification and localization at the query level. During training, however, each ground-truth object is still matched to a single object query. Compared with these methods, SPMR applies one-to-one supervision to class-agnostic proposal generation and uses multiple localized proposals for proposal-level fine-grained recognition.

\begin{figure*}[!t]
  \begin{center}
    \includegraphics[width=\textwidth]{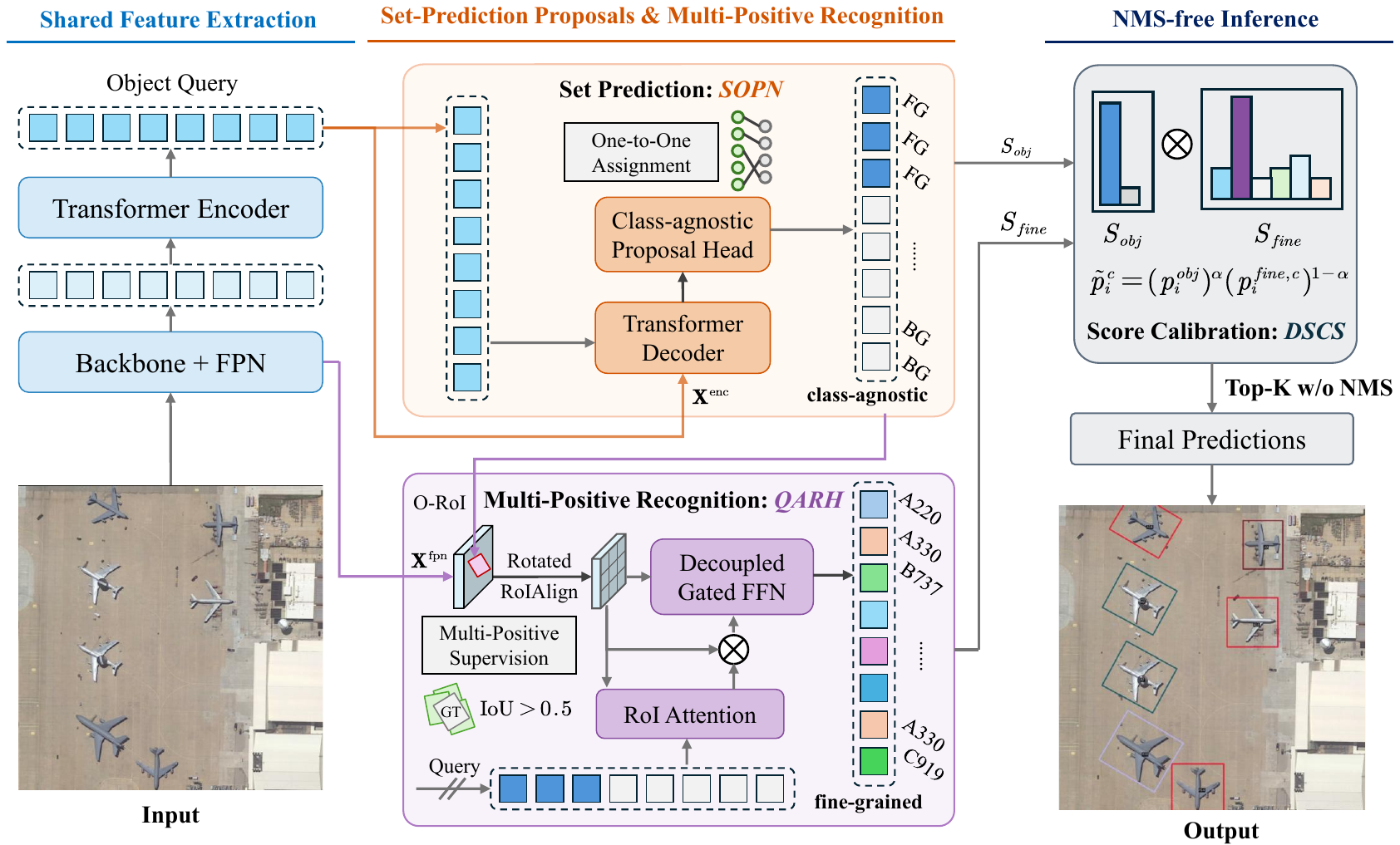}
  \end{center}
  \caption{Overview of the proposed SPMR. Given an input image, the backbone and Feature Pyramid Network (FPN) extract multi-scale features, which are encoded by a Transformer encoder. SOPN generates class-agnostic oriented proposals under one-to-one supervision, and QARH performs proposal-level fine-grained recognition on these proposals. During inference, DSCS fuses the objectness score and fine-grained class score to produce NMS-free predictions. The training-only DPAS branch is omitted for clarity.}
  \label{fig:spmr_overview}
\end{figure*}

\section{Method} \label{sec:method}
\subsection{Overview of SPMR Framework}
SPMR is designed to assign complementary roles to sparse proposal generation and fine-grained recognition in NMS-free FGOD. As illustrated in \autoref{fig:spmr_overview}, the backbone and FPN first extract multi-scale features $\mathbf{X}^{\mathrm{fpn}}$, which are further encoded by a Transformer encoder into $\mathbf{X}^{\mathrm{enc}}$. Based on these features, SOPN generates a sparse proposal set and estimates proposal objectness, while QARH integrates query and RoI features for fine-grained recognition and box refinement. During training, SOPN predictions are matched one-to-one with ground-truth objects, whereas multiple localized proposals are used as positives for QARH recognition. During inference, DSCS combines SOPN objectness and QARH class scores to rank the refined proposals without NMS.

\subsection{Sparse Oriented Proposal Network} \label{sec:SOPN}
SOPN is built on a customized oriented variant of DINO~\cite{dino}, denoted as DINO-O. To adapt the original horizontal detector to oriented detection, we make two oriented-detection adaptations. Specifically, we normalize angles to the range $[-\pi/2, \pi/2]$ and integrate rotational deformable attention~\cite{arsdetr} to mitigate feature misalignment caused by object rotation. In addition to these adaptations, a critical challenge in training query-based oriented detectors is the numerical disparity between bounding box coordinates and rotation angles, which can cause angular errors to dominate the matching cost, leading to suboptimal bipartite matching. To address this issue without introducing complex angle encoding schemes~\cite{arsdetr, rhino, qedetr}, we adopt a concise angle-weighted $\mathcal{L}_{1}$ loss. The loss is formulated as:
\begin{equation}
  \mathcal{L}_{\mathrm{bbox}} = \mathcal{L}_{\mathrm{xywh}} + \lambda_{\theta} \mathcal{L}_{\theta},
  \label{eq:sopn_bbox}
\end{equation}
where $\mathcal{L}_{\mathrm{xywh}}$ denotes the $\mathcal{L}_{1}$ loss for center coordinates and box dimensions, and $\mathcal{L}_{\theta}$ denotes the $\mathcal{L}_{1}$ loss for orientation. We use the same angular balancing in both the matching cost and the regression loss. Specifically, the Hungarian matching cost is composed of focal loss cost, rotated $\mathcal{L}_{1}$ cost, and rotated IoU cost with weights of $2.0$, $5.0$, and $2.0$, respectively, where the angular term in the rotated $\mathcal{L}_{1}$ cost is weighted by $\lambda_{\theta}=0.2$.

To facilitate coordination with the subsequent recognition stage, SOPN decouples the features used for sparse proposal generation from those used for proposal-level recognition. Since the global features from Transformer encoders are well-suited for localization but may lack the local details required for fine-grained recognition~\cite{petdet, drnet}, we retain the FPN~\cite{fpn} to provide multi-scale representations. Because the refinement head already supplies proposal-level feature interaction for recognition, SOPN does not need a deep encoder to carry all semantic discrimination by itself, allowing us to reduce the Transformer encoder depth from six layers to two and use the saved computation for QARH and the FPN features.

Importantly, SOPN operates under a class-agnostic training paradigm. By mapping all fine-grained classes to a single foreground class, the network focuses exclusively on distinguishing objects from the background and regressing precise boundaries. This strategy helps reduce the interference of fine-grained semantic confusion in localization learning. During training, we employ focal loss for classification and a combination of the proposed angle-weighted $\mathcal{L}_{1}$ loss and rotated IoU loss for regression. These set-prediction losses are applied at each decoder layer, and contrastive denoising is adopted following DINO~\cite{dino}.

\subsection{Query-Aware Refinement Head} \label{sec:qarh}
The Query-Aware Refinement Head (QARH) performs fine-grained recognition on the sparse proposals generated by SOPN. It takes detached SOPN proposal boxes and query features as proposal-level inputs, extracts local RoI features from the FPN, and integrates them with global query semantics for recognition. This detachment design prevents the fine-grained recognition loss from directly updating SOPN proposal boxes and query features, while still allowing QARH and shared image features to learn proposal-level fine-grained discrimination.

\noindent\textbf{Query-Aware Decoupled Refinement.}
QARH extracts complementary features from SOPN query features and RoI features, capturing both global semantics and local details, and dynamically filters the fused representation for classification and regression. For the $i$-th proposal, let $\mathbf{q}_i$ denote its query feature from SOPN and $\mathbf{F}_i^{\mathrm{roi}}$ denote its RoI feature map extracted via RoIAlign from FPN. QARH derives two complementary feature vectors:
\begin{equation}
  \begin{aligned}
    \mathbf{v}_i^{\mathrm{qa}}  & = \operatorname{RoIAttn}(\mathbf{q}_i, \mathbf{F}_i^{\mathrm{roi}}),  \\
    \mathbf{v}_i^{\mathrm{roi}} & = \mathbf{W}_{p}(\operatorname{Flatten}(\mathbf{F}_i^{\mathrm{roi}})),
  \end{aligned}
\end{equation}
where $\operatorname{RoIAttn}$ implements cross-attention to extract query-aware semantic context, inspired by HPR~\cite{hpr}, and $\mathbf{W}_{p}$ is a linear projection layer that transforms spatial features. The query-aware vector $\mathbf{v}_i^{\mathrm{qa}}$ encodes global semantic context, while the RoI vector $\mathbf{v}_i^{\mathrm{roi}}$ preserves spatial details.

\begin{figure}[!t]
  \begin{center}
    \includegraphics[width=0.43\textwidth]{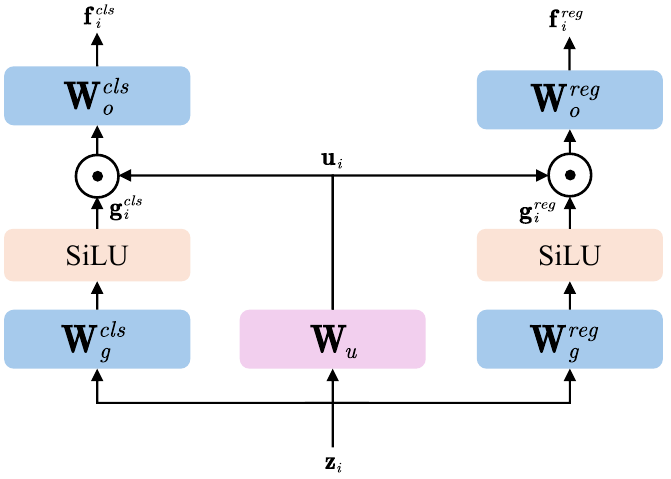}
  \end{center}
  \caption{Decoupled Gated FFN for task-specific refinement. The classification and regression branches apply separate gates to the shared content representation to obtain task-specific embeddings.}
  \label{fig:ffn}
\end{figure}

For task-specific refinement, we use a Decoupled Gated FFN~\cite{gluvariants} to modulate the shared representation separately for classification and regression, as illustrated in \autoref{fig:ffn}. The RoI feature $\mathbf{v}_i^{\mathrm{roi}}$ and query-aware feature $\mathbf{v}_i^{\mathrm{qa}}$ are concatenated as $\mathbf{z}_i = [\mathbf{v}_i^{\mathrm{roi}}; \mathbf{v}_i^{\mathrm{qa}}]$. The concatenated representation is projected into two task-specific gates and a shared feature representation:
\begin{equation}
  \begin{aligned}
    \mathbf{g}_{i}^{cls} & = \sigma(\mathbf{W}_{g}^{cls} \mathbf{z}_i), \\
    \mathbf{g}_{i}^{reg} & = \sigma(\mathbf{W}_{g}^{reg} \mathbf{z}_i), \\
    \mathbf{u}_i         & = \mathbf{W}_{u} \mathbf{z}_i,
  \end{aligned}
\end{equation}
where $\sigma(\cdot)$ denotes the SiLU activation, and $\mathbf{W}_{g}^{cls}$, $\mathbf{W}_{g}^{reg}$, and $\mathbf{W}_{u}$ are learnable projections. The task-specific gates then modulate the shared representation $\mathbf{u}_i$:
\begin{equation}
  \begin{aligned}
    \mathbf{f}_{i}^{cls} & = \mathbf{W}_{o}^{cls}(\mathbf{g}_{i}^{cls} \odot \mathbf{u}_i), \\
    \mathbf{f}_{i}^{reg} & = \mathbf{W}_{o}^{reg}(\mathbf{g}_{i}^{reg} \odot \mathbf{u}_i),
  \end{aligned}
\end{equation}
where $\mathbf{W}_{o}^{cls}, \mathbf{W}_{o}^{reg}$ are projection matrices, and $\odot$ denotes element-wise multiplication. The task-specific gates ($\mathbf{g}_{i}^{cls}$ and $\mathbf{g}_{i}^{reg}$) enable classification and regression tasks to selectively attend to task-relevant information from the shared representation $\mathbf{u}_i$. The resulting task-specific embeddings are used to predict fine-grained class scores and class-agnostic box refinements, with the latter optimized by a smooth $\mathcal{L}_{1}$ loss on positive proposals. Thus, QARH improves fine-grained recognition while still refining the sparse SOPN proposals.

\noindent\textbf{Multi-Positive Recognition.}
During training, QARH retains the full sparse SOPN proposal set rather than randomly sampling a subset of positive and negative proposals, allowing multiple proposals around the same object to contribute to recognition. Following standard RoI assignment~\cite{orcnn}, each pre-refinement proposal is assigned to the ground-truth box with the highest rotated IoU. A proposal is labeled positive and assigned the corresponding fine-grained class when this maximum IoU is at least $0.5$. All other proposals are labeled background. The same assignment rule is applied to denoising and DPAS proposals.

We further weight each proposal by its objectness and localization quality using an exponential scaling rule. This weighting design encourages QARH to focus on proposals that are consistently favored by SOPN and accurately localized after refinement. In this formulation, the SOPN objectness score provides a query-aware cross-stage consistency prior, while the IoU term emphasizes high-quality refined proposals. Let $C$ denote the number of foreground classes. For the $i$-th proposal, let $y_i \in \{0, \ldots, C-1, \mathrm{bg}\}$ denote its ground-truth label and $\mathbf{p}_i^{\mathrm{fine}}$ be the probability distribution predicted by QARH over all classes. The loss is formulated as:
\begin{equation}
  \mathcal{L}_{\mathrm{MPR}} = \sum_{i} w_i \, \operatorname{CE}(\mathbf{p}_i^{\mathrm{fine}}, y_i),
  \label{eq:mpr_loss}
\end{equation}
where CE denotes cross-entropy. To better formulate the sample weights, we first introduce a quality term that combines localization and objectness:
\begin{equation}
  \tau_i = p_i^{\mathrm{obj}} \cdot \operatorname{IoU}_i,
\end{equation}
where $p_i^{\mathrm{obj}}$ represents the objectness confidence predicted by SOPN, and $\operatorname{IoU}_i$ is the rotated IoU between the $i$-th QARH-refined box and its assigned ground truth. Both $p_i^{\mathrm{obj}}$ and $\operatorname{IoU}_i$ are detached before computing the classification-loss weight. The sample weight $w_i$ is then computed as:
\begin{equation}
  w_i = \begin{cases}
    \tau_i \exp(\kappa \tau_i), & \text{if } y_i \neq \mathrm{bg}, \\
    p_i^{\mathrm{obj}},         & \text{if } y_i = \mathrm{bg}
  \end{cases},
\end{equation}
where $\kappa$ controls the strength of the exponential scaling effect and is fixed to $2.0$ in our experiments. For positive proposals, the exponential form amplifies high-quality samples while suppressing proposals with weak objectness or poor localization, reducing the risk that ambiguous low-quality RoIs dominate fine-grained recognition learning. For negative proposals, the weight is directly modulated by the objectness score $p_i^{\mathrm{obj}}$, encouraging QARH to focus more on hard negatives that SOPN considers object-like and to allocate less learning emphasis to easy background samples. Consequently, MPR integrates multi-positive supervision with quality-aware sample weighting to enhance fine-grained recognition within the sparse proposal set.

\subsection{Dual-Path Auxiliary Supervision} \label{sec:dpas}
In sparse set-prediction detectors, the training signals are limited by the fixed query set, which becomes more restrictive in FGOD because learning subtle inter-class differences requires sufficiently localized positive candidates. SPMR uses DPAS as a training-only mechanism that provides additional localized proposal samples while leaving the inference path unchanged.

On the query side, DPAS adds an auxiliary encoder head that applies one-to-many matching only to auxiliary encoder candidates. This auxiliary head provides denser object-level supervision, while the main SOPN branch remains trained with one-to-one set prediction. The auxiliary encoder head selects $300$ high-scoring encoder proposals to initialize an auxiliary query group. For this auxiliary supervision, each ground truth is repeated $3$ times, and the auxiliary branch follows the same class-agnostic objectness setting, matching weights, and loss weights as SOPN. These auxiliary queries are then refined by the SOPN decoder and trained with the same one-to-one set-prediction objective as the main queries. In this way, densely supervised auxiliary candidates are filtered under the sparse SOPN objective, helping SOPN learn more reliable query selection.

On the recognition side, the refined auxiliary queries are also used by QARH during training. They are converted into auxiliary RoIs and forwarded to the fine-grained recognition head, where they reuse the same proposal-level multi-positive supervision and quality-weighted classification loss as the main SOPN proposals. This gives QARH more localized training cases while keeping the main inference path unchanged. During inference, SPMR uses only the main SOPN-QARH path, and the auxiliary encoder head and auxiliary query group are omitted.

With DPAS, the training objective is written as:
\begin{equation}
  \mathcal{L}_{\mathrm{total}} = \mathcal{L}_{\mathrm{SOPN}} + \mathcal{L}_{\mathrm{QARH}} + \mathcal{L}_{\mathrm{DPAS}},
\end{equation}
where $\mathcal{L}_{\mathrm{SOPN}}$ and $\mathcal{L}_{\mathrm{QARH}}$ are the losses of the main SOPN and QARH branches, and $\mathcal{L}_{\mathrm{DPAS}}$ collects the auxiliary losses from the query-side and recognition-side paths.

\subsection{Dual-Stage Consensus Scoring} \label{sec:dscs}
During inference, Dual-Stage Consensus Scoring (DSCS) combines the objectness confidence from SOPN and the fine-grained classification probability from QARH for each proposal before final ranking.

We denote SOPN objectness as $S_{\mathrm{obj}} = p_i^{\mathrm{obj}}$, where $p_i^{\mathrm{obj}}$ is the SOPN foreground probability, and QARH softmax outputs as $S_{\mathrm{fine}} = \mathbf{p}_i^{\mathrm{fine}}$. Let $\mathbf{p}_i^{\mathrm{fine}} = [p_i^{\mathrm{fine},0}, \ldots, p_i^{\mathrm{fine},C-1}, p_i^{\mathrm{fine},bg}]$ be the fine-grained classification probability distribution predicted by QARH. For each foreground class $c \in \{0,1,\ldots,C-1\}$, a high score should reflect both class-agnostic objectness and class-specific recognition evidence. To balance the contributions of the SOPN objectness score and QARH fine-grained class score, we use a weighted geometric score:
\begin{equation}
  \tilde{p}_i^c = (p_i^{\mathrm{obj}})^{\alpha} \cdot (p_i^{\mathrm{fine},c})^{1-\alpha}.
  \label{eq:dscs_geo}
\end{equation}

Here, $\alpha \in [0,1]$ controls the relative contributions of the SOPN objectness score and QARH fine-grained class score. The QARH background probability is used in the standard $(C+1)$-way training loss. At inference, DSCS derives the background score from SOPN objectness:
\begin{equation}
  \tilde{p}_i^{bg} = 1 - p_i^{\mathrm{obj}}.
  \label{eq:dscs_background}
\end{equation}

The fused foreground and background scores form the unnormalized vector $\tilde{\mathbf{p}}_i = [\tilde{p}_i^0, \ldots, \tilde{p}_i^{C-1}, \tilde{p}_i^{bg}]$. We apply $\ell_1$ normalization to obtain the final class scores.
\begin{equation}
  \mathbf{p}_i^{\mathrm{final}} = \frac{\tilde{\mathbf{p}}_i}{\lVert \tilde{\mathbf{p}}_i \rVert_1}.
  \label{eq:dscs_final}
\end{equation}
The resulting $\mathbf{p}_i^{\mathrm{final}}$ is a $(C+1)$-dimensional score distribution whose entries sum to one. At inference, its $C$ foreground components are used to rank the refined QARH predictions globally, producing the final detections without NMS.

\section{Experimental Results and Analysis} \label{sec:result}
To validate the effectiveness of SPMR, we conduct experiments on MAR20, ShipRSImageNet, and FAIR1M. We first compare SPMR with state-of-the-art methods, and then perform ablation studies to analyze the contribution of each component.

\subsection{Datasets}
We employ three FGOD datasets, including MAR20~\cite{mar20}, ShipRSImageNet~\cite{shiprs}, and FAIR1M~\cite{fair1m}, to validate the proposed method. We use the MAR20 dataset as the primary benchmark for our experimental analysis.

\subsubsection{MAR20 Dataset}
MAR20~\cite{mar20} is a remote sensing dataset for military aircraft recognition. All instances in the MAR20 dataset are annotated at the specific model level, covering 20 aircraft types. The image sizes are primarily $800 \times 800$ pixels. MAR20 contains a total of 3,842 images, with 1,331 for training and 2,511 for testing. The dataset includes 22,341 aircraft instances, all of which are annotated with both horizontal bounding boxes (HBB) and oriented bounding boxes (OBB).

\subsubsection{ShipRSImageNet Dataset}
ShipRSImageNet~\cite{shiprs} is a large-scale dataset for fine-grained ship detection. The image sizes vary from $930 \times 930$ to $1024 \times 1024$ pixels. ShipRSImageNet contains 3,435 images, with 2,198 for training, 550 for validation, and 687 for testing, totaling 17,573 ship instances. All instances are annotated with HBB, OBB, and polygon masks. All objects are annotated on a coarse-to-fine scale from level 0 to 3. We conduct our experiments using the level-3 classification, where ships are divided into 50 fine-grained types. Since the annotations for the test set are not public, we train on the official training set and test on the validation set.

\subsubsection{FAIR1M Dataset}
FAIR1M~\cite{fair1m} includes five coarse-grained categories (Airplane, Ship, Vehicle, Court, Road) and 37 fine-grained categories. The data were collected from various sensors and platforms, with spatial resolutions ranging from 0.3--0.8~m. The image sizes range from $1000 \times 1000$ to $10{,}000 \times 10{,}000$ pixels. FAIR1M-v1.0 has 16,488 images in the training set and 8,137 in the test set. FAIR1M-v2.0 expands the test set to 18,021 images and adds a validation set of 8,287 images. All instances are annotated with OBB. Following previous studies, we combine the training and validation sets for training on FAIR1M-v2.0.

\subsection{Implementation Details}
All experiments are implemented based on the MMRotate 1.x toolbox~\cite{mmrotate} with four NVIDIA GeForce RTX 2080 Ti GPUs. Unless otherwise specified, we employ ResNet50~\cite{resnet} as the backbone. The total batch size is set to $8$, and for all query-based detectors, including our proposed SPMR, the number of object queries is fixed at $300$. Unless otherwise noted, all comparison results are obtained using our implementation.

\noindent\textbf{Data Preprocessing and Augmentation.}
For the MAR20 and ShipRSImageNet datasets, images are resized to fit within $800 \times 800$ and $1024 \times 1024$ patches, respectively, while preserving the aspect ratio. We apply horizontal, vertical, or diagonal flipping with a probability of $0.25$ for each direction. For FAIR1M, we crop the original images into $1024 \times 1024$ patches with an overlap of $200$ pixels in the single-scale setting and apply the same flipping policy without random rotation. For multi-scale training, we follow QEDetr~\cite{qedetr}, increase the overlap to $300$ pixels, use scale factors of $[0.5, 1.0, 1.5]$, and apply random rotation with a probability of $0.5$.

\noindent\textbf{Optimization and Training Schedules.}
We use architecture-specific optimization strategies based on the recommended configurations for the respective methods. For CNN-based methods, we use SGD with momentum $0.9$ and weight decay $1 \times 10^{-4}$. The initial learning rates are $1 \times 10^{-2}$ and $2 \times 10^{-2}$ for one-stage and two-stage methods, respectively. For SPMR and the reimplemented query-based methods, we use AdamW with initial learning rates of $2 \times 10^{-4}$ for SPMR and Deformable-DETR-O and $1 \times 10^{-4}$ for the other query-based methods. The training schedules are adjusted according to the dataset scale. For MAR20 and ShipRSImageNet, models are trained for $36$ epochs, with the learning rate decayed by a factor of $10$ at epochs $24$ and $33$ for CNN-based models, and at epoch $30$ for query-based models. For the FAIR1M dataset, we adopt a $12$-epoch schedule, where the learning rate is decayed at epochs $8$ and $11$ for CNN-based models, and at epoch $11$ for query-based models.

\subsection{Evaluation Metric}
We use mean Average Precision (AP) as the evaluation metric for FGOD. Due to the emphasis on recognition performance in FGOD, AP refers to $\mathrm{AP_{50}}$ by default, unless otherwise specified. This follows the convention of previous FGOD studies. For FAIR1M, since the labels are not public, the online evaluation server reports only $\mathrm{AP_{50}}$. This metric is calculated over 34 fine-grained classes, excluding 'Other Airplane', 'Other Ship', and 'Other Vehicle'. For MAR20 and ShipRSImageNet, we compute $\mathrm{AP_{50}}$ locally. We also report $\mathrm{AP_{75}}$ and $\mathrm{AP_{50:95}}$ to assess performance under stricter localization requirements.

\subsection{Comparison with State-of-the-Art Methods}
To objectively evaluate the performance of SPMR, we compare it with mainstream FGOD paradigms, including traditional two-stage detectors, one-stage detectors, and NMS-free set-prediction detectors. The experimental results show that SPMR achieves competitive performance on multiple datasets without requiring NMS.

\subsubsection{Quantitative and Qualitative Results on MAR20}
\begin{table}[!t]
  \centering
  \caption{Comparison results on MAR20}
  \begin{tblr}{
    colsep = 5.4pt,
    column{2-5} = {c},
    cell{2}{1} = {c=4}{l},
    cell{7}{1} = {c=4}{l},
    cell{14}{1} = {c=4}{l},
    hline{3} = {-}{},
    hline{7,8} = {-}{},
    hline{14,15} = {-}{},
      }
    \toprule
    Method                                 & Venue & $\mathrm{AP_{50}}$ & $\mathrm{AP_{75}}$ & $\mathrm{AP_{50:95}}$ \\
    \midrule
    \textit{One-stage methods:}                                                                                          \\
    RetinaNet-O~\cite{retinanet}           & ICCV2017  & 67.35              & 34.71              & 36.44                 \\
    \text{FCOS-O}~\cite{fcos}              & ICCV2019  & 70.53              & 55.90              & 47.08                 \\
    R$^3$Det~\cite{r3det}                  & AAAI2021  & 64.24              & 33.69              & 35.56                 \\
    S$^2$ANet~\cite{s2anet}                & TGRS2022  & 77.38              & 51.06              & 46.88                 \\
    \textit{Two-stage methods:}                                                                                          \\
    Faster R-CNN-O~\cite{faster}           & NIPS2015  & 80.30              & 57.11              & 50.99                 \\
    RoI Tran~\cite{roitrans}               & CVPR2019  & 82.15              & 73.06              & 58.40                 \\
    Oriented R-CNN~\cite{orcnn}            & ICCV2021  & 82.71              & 72.15              & 58.25                 \\
    PCLDet~\cite{pcldet}                   & TGRS2023  & 82.02              & 71.82              & 57.57                 \\
    PETDet~\cite{petdet}                   & TGRS2024  & 85.06              & 76.67              & 60.84                 \\
    DRNet~\cite{drnet}                     & TCSVT2025 & 82.51              & 73.61              & 58.79                 \\
    \textit{NMS-free methods:}                                                                                           \\
    \text{Deformable-DETR-O}~\cite{deform} & ICLR2021  & 76.10              & 46.83              & 44.74                 \\
    DINO-O~\cite{dino}                     & ICLR2023  & 75.36              & 62.11              & 50.65                 \\
    ARS-DETR~\cite{arsdetr}                & TGRS2024  & 76.28              & 63.47              & 51.13                 \\
    OrientedFormer~\cite{oformer}          & TGRS2024  & 80.95              & 62.38              & 53.30                 \\
    RHINO~\cite{rhino}                     & WACV2025  & 75.33              & 52.40              & 47.17                 \\
    \textbf{SPMR (Ours)}                   & ---       & \textbf{85.82}     & \textbf{77.54}     & \textbf{61.44}        \\
    \bottomrule
  \end{tblr}
  \label{tab:mar20}
\end{table}

As shown in \autoref{tab:mar20}, two-stage detectors generally outperform one-stage and set-prediction methods, confirming that structural decoupling benefits FGOD. SPMR achieves 85.82 AP, improving over PETDet by 0.76 AP and DINO-O by 10.46 AP. This result supports the proposed role-separated design, which combines NMS-free proposal generation with proposal-level fine-grained recognition.

\begin{table}[!t]
  \centering
  \caption{Per-class $\mathrm{AP}_{50}$ comparison with representative NMS-free methods on MAR20}
  \begin{tblr}{
    stretch=1.0,
    cell{1}{1} = {r=2}{},
    row{1} = {belowsep=0pt},
    colsep=4.5pt,
    colspec={X[0.65,c]X[1.1,c]X[c]X[c]X[c]X[1.1,c]X[c]},
    hline{23} = {-}{}
    }
    \toprule
    Type & \text{D-DETR}  & DINO        & ARS            & RHINO        & OFormer        & SPMR           \\
         & \cite{deform}  & \cite{dino} & \cite{arsdetr} & \cite{rhino} & \cite{oformer} & --             \\
    \midrule
    A1   & 77.75          & 69.34       & 71.95          & 74.10        & 79.56          & \textbf{88.67} \\
    A2   & 87.36          & 82.66       & 83.47          & 83.39        & 83.36          & \textbf{90.28} \\
    A3   & 85.08          & 82.70       & 81.76          & 83.35        & 86.28          & \textbf{89.05} \\
    A4   & 80.13          & 77.82       & 80.83          & 75.91        & 76.66          & \textbf{85.19} \\
    A5   & 54.53          & 63.71       & 52.04          & 61.81        & 70.93          & \textbf{76.95} \\
    A6   & \textbf{90.88} & 88.68       & 85.73          & 86.13        & 90.56          & 90.82          \\
    A7   & 87.01          & 79.51       & 82.13          & 79.97        & 88.45          & \textbf{90.19} \\
    A8   & 87.01          & 82.10       & 82.08          & 78.16        & 86.88          & \textbf{88.71} \\
    A9   & 88.48          & 84.24       & 86.03          & 83.19        & 88.07          & \textbf{90.86} \\
    A10  & 88.23          & 88.36       & 85.06          & 89.85        & 90.73          & \textbf{90.86} \\
    A11  & 81.43          & 73.38       & 77.18          & 74.17        & 83.35          & \textbf{86.04} \\
    A12  & 86.18          & 78.52       & 84.62          & 78.34        & 85.44          & \textbf{88.00} \\
    A13  & 58.60          & 68.98       & 70.00          & 63.24        & 66.75          & \textbf{77.87} \\
    A14  & 82.39          & 79.66       & 79.70          & 78.33        & 84.16          & \textbf{89.29} \\
    A15  & 45.13          & 47.81       & 56.78          & 53.43        & 59.24          & \textbf{67.05} \\
    A16  & 51.41          & 73.32       & 71.55          & 77.36        & \textbf{86.99} & 84.99          \\
    A17  & 90.10          & 88.65       & 88.20          & 87.47        & 90.08          & \textbf{90.35} \\
    A18  & 44.26          & 50.81       & 57.27          & 51.77        & 65.39          & \textbf{84.05} \\
    A19  & 82.22          & 74.66       & 76.10          & 73.05        & 79.59          & \textbf{86.18} \\
    A20  & 73.84          & 72.21       & 73.12          & 73.65        & 76.60          & \textbf{81.05} \\
    AP   & 76.10          & 75.36       & 76.28          & 75.33        & 80.95          & \textbf{85.82} \\
    \bottomrule
  \end{tblr}
  \label{tab:mar20_sub}
\end{table}

\begin{figure}[!t]
  \begin{center}
    \includegraphics[width=\linewidth]{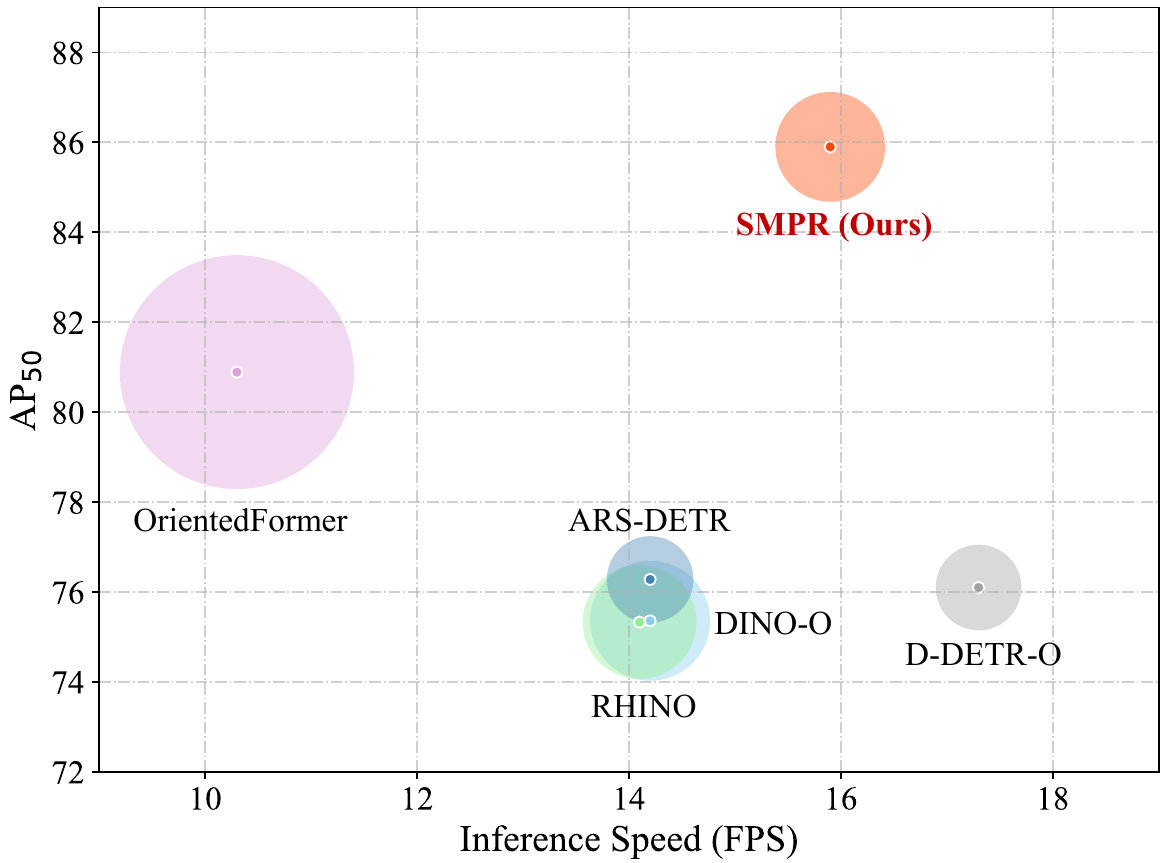}
  \end{center}
  \caption{Speed-accuracy trade-off comparison with representative NMS-free detectors on the MAR20 dataset. The x-axis and y-axis denote inference speed (FPS) and detection accuracy ($\mathrm{AP}_{50}$), respectively. The bubble size represents the computational cost (FLOPs). SPMR shows a favorable balance between fine-grained recognition accuracy and inference efficiency.}
  \label{fig:speed_accuracy}
\end{figure}

\autoref{fig:speed_accuracy} illustrates the performance-efficiency trade-off, with all speeds measured on a single RTX 2080 Ti GPU under the same input setting. SPMR appears in the region with both high accuracy and high inference speed. Specifically, SPMR outperforms OrientedFormer by +4.87 AP while running $1.5\times$ faster ($15.9$ vs. $10.3$ FPS) with lower computational cost ($143.3$ GFLOPs vs. $324.6$ GFLOPs). This speed advantage mainly comes from the lighter set-prediction stage, where the Transformer encoder is reduced to two layers and inference does not include NMS. The reduced encoder offsets much of the additional cost introduced by QARH, while QARH operates only on a sparse proposal set rather than dense feature locations. Compared with the fastest compared method, D-DETR-O, SPMR achieves a +9.72 AP gain with a moderate speed cost.

\begin{figure*}[!t]
  \centering
  \begin{minipage}[c]{.02\textwidth}
    \rotatebox{90}{(a) gt}
  \end{minipage}
  \begin{minipage}[c]{.157\textwidth}
    \centering
    \includegraphics[width=\textwidth]{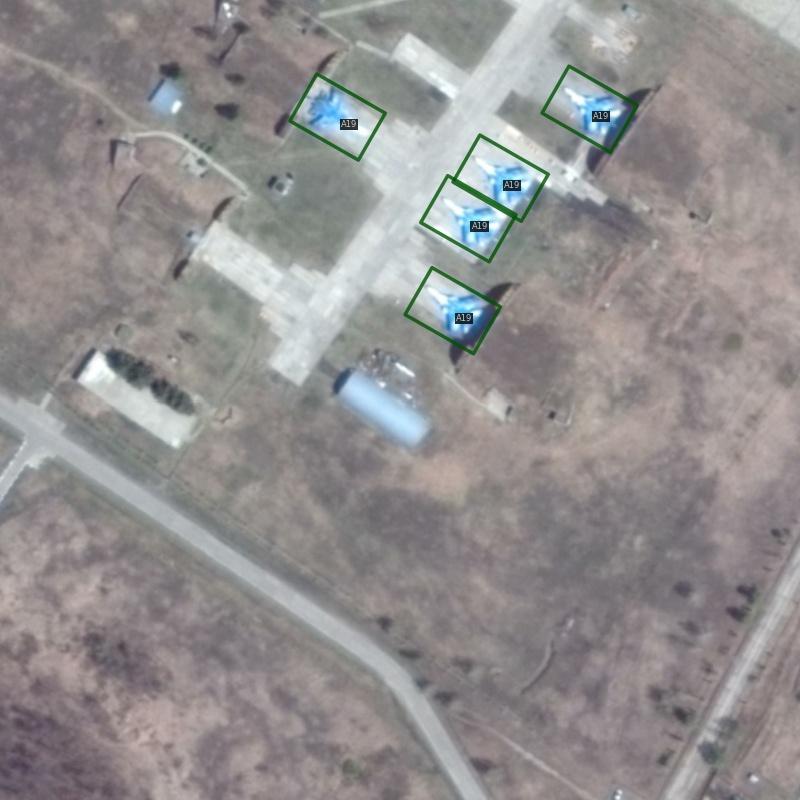}
  \end{minipage}
  \begin{minipage}[c]{.157\textwidth}
    \centering
    \includegraphics[width=\textwidth]{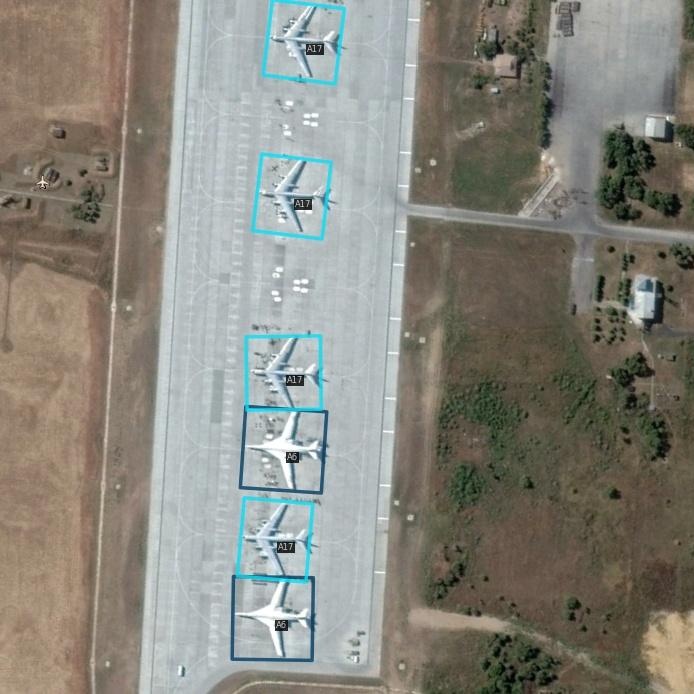}
  \end{minipage}
  \begin{minipage}[c]{.157\textwidth}
    \centering
    \includegraphics[width=\textwidth]{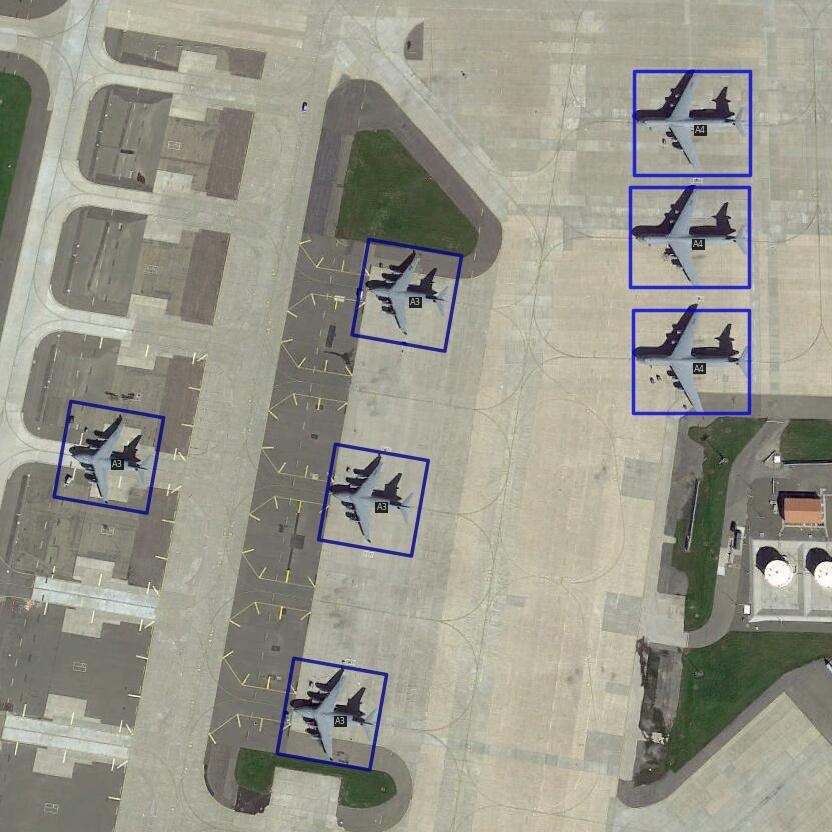}
  \end{minipage}
  \begin{minipage}[c]{.157\textwidth}
    \centering
    \includegraphics[width=\textwidth]{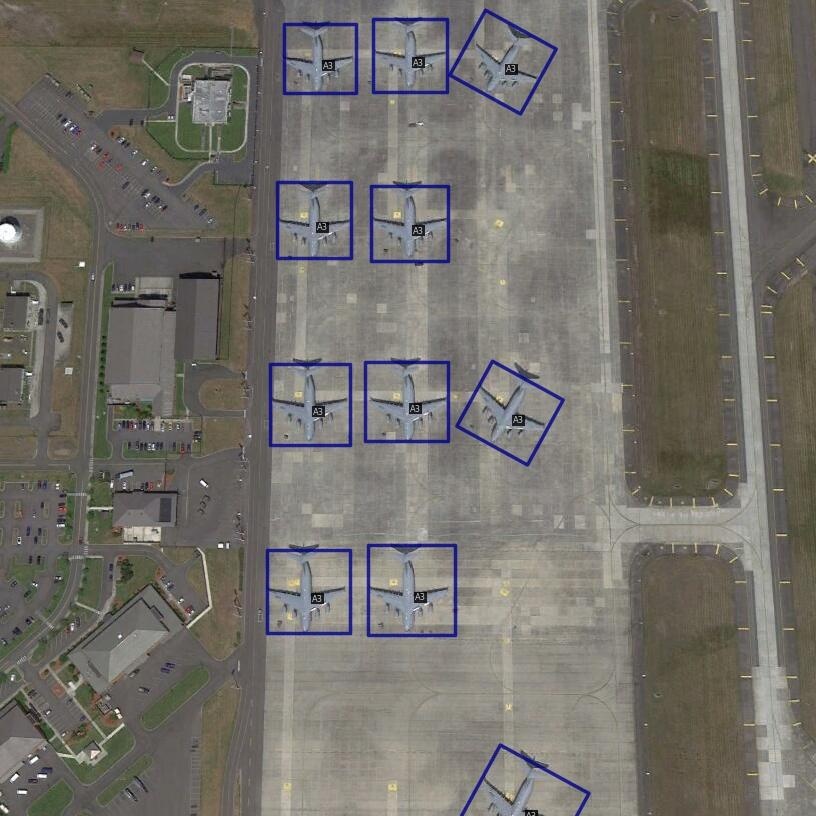}
  \end{minipage}
  \begin{minipage}[c]{.157\textwidth}
    \centering
    \includegraphics[width=\textwidth]{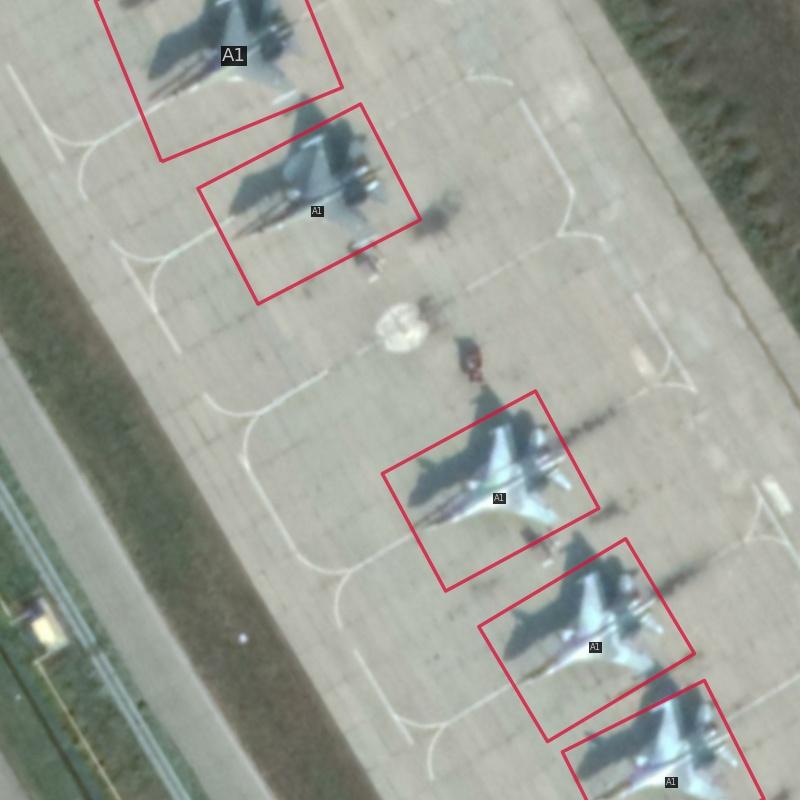}
  \end{minipage}
  \begin{minipage}[c]{.157\textwidth}
    \centering
    \includegraphics[width=\textwidth]{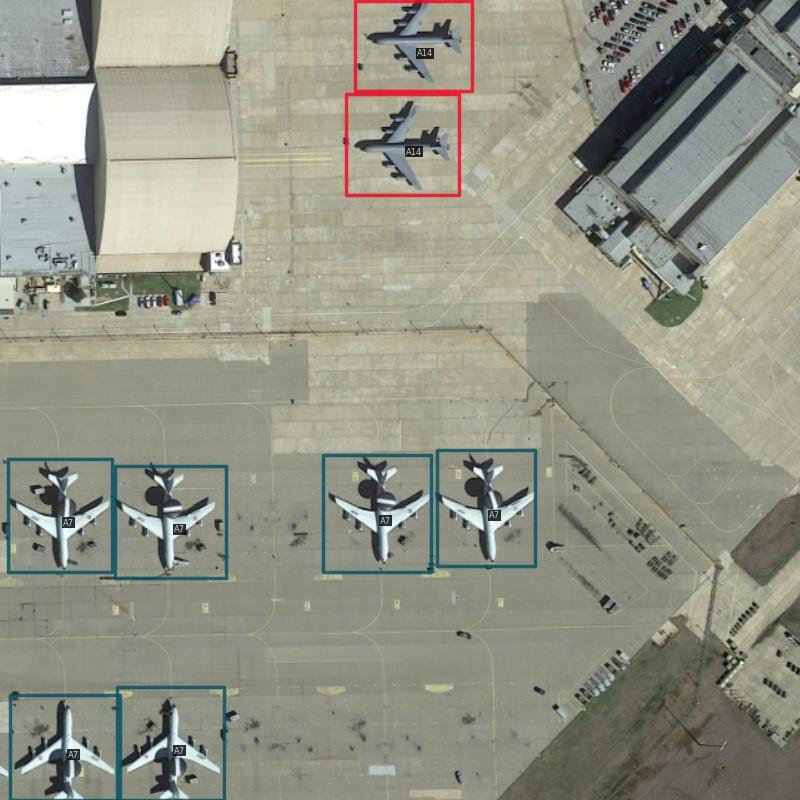}
  \end{minipage}

  \vspace{1pt}
  \centering
  \begin{minipage}[c]{.02\textwidth}
    \rotatebox{90}{(b) baseline}
  \end{minipage}
  \begin{minipage}[c]{.157\textwidth}
    \centering
    \includegraphics[width=\textwidth]{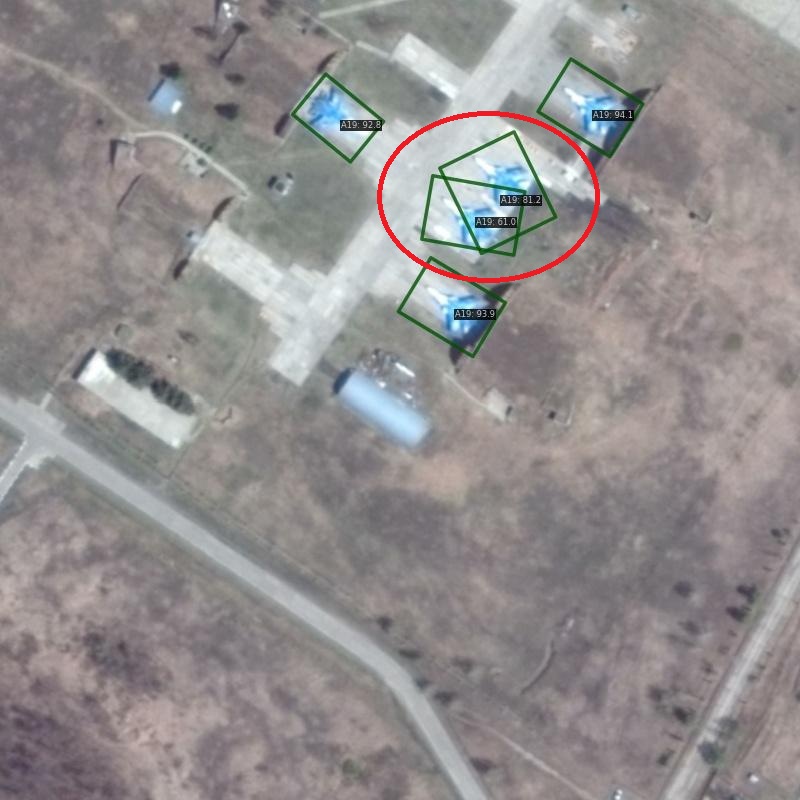}
  \end{minipage}
  \begin{minipage}[c]{.157\textwidth}
    \centering
    \includegraphics[width=\textwidth]{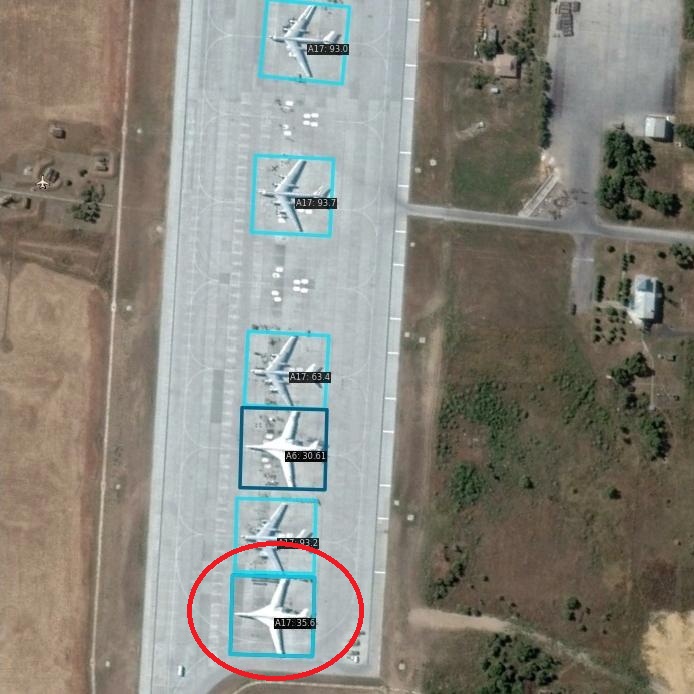}
  \end{minipage}
  \begin{minipage}[c]{.157\textwidth}
    \centering
    \includegraphics[width=\textwidth]{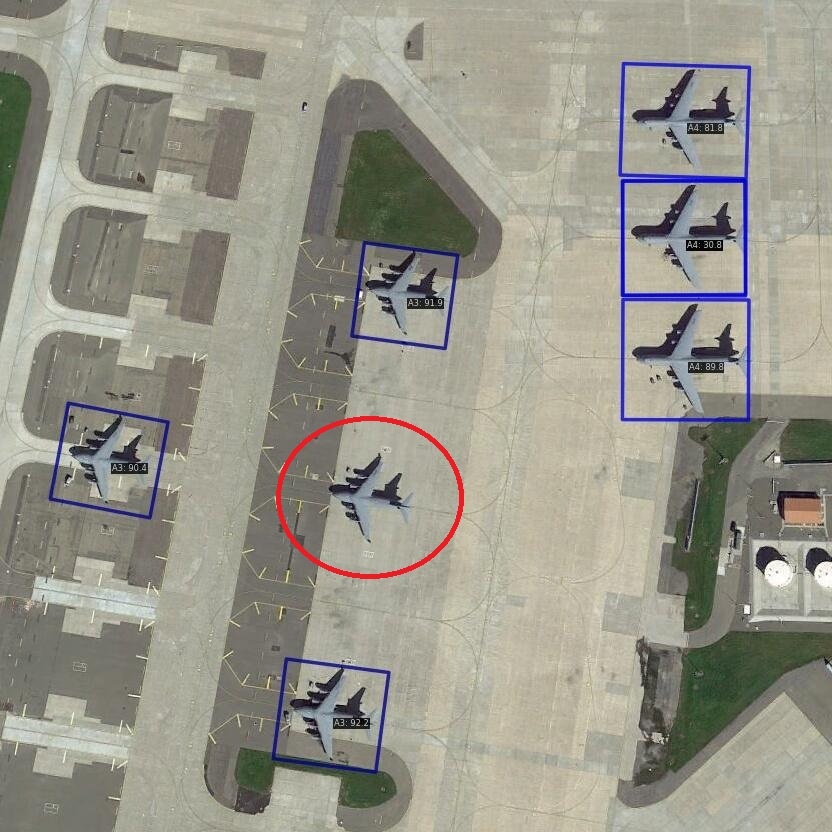}
  \end{minipage}
  \begin{minipage}[c]{.157\textwidth}
    \centering
    \includegraphics[width=\textwidth]{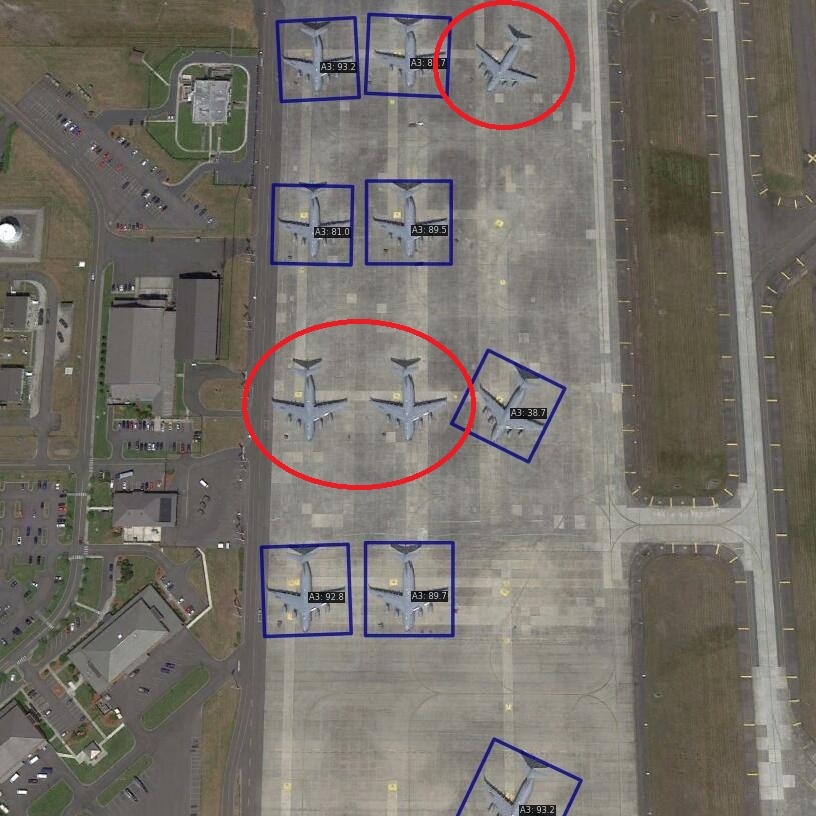}
  \end{minipage}
  \begin{minipage}[c]{.157\textwidth}
    \centering
    \includegraphics[width=\textwidth]{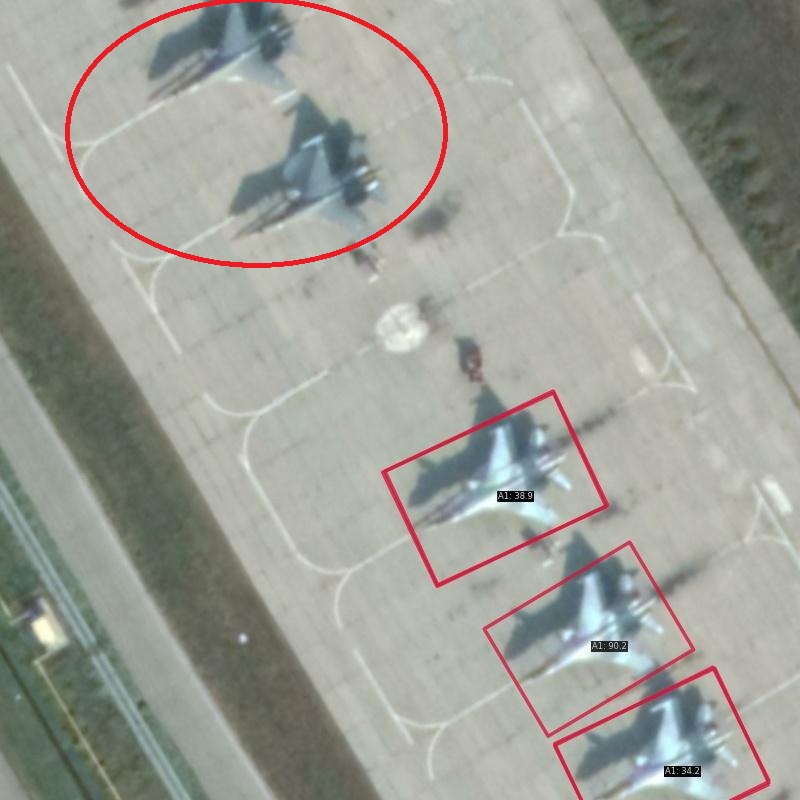}
  \end{minipage}
  \begin{minipage}[c]{.157\textwidth}
    \centering
    \includegraphics[width=\textwidth]{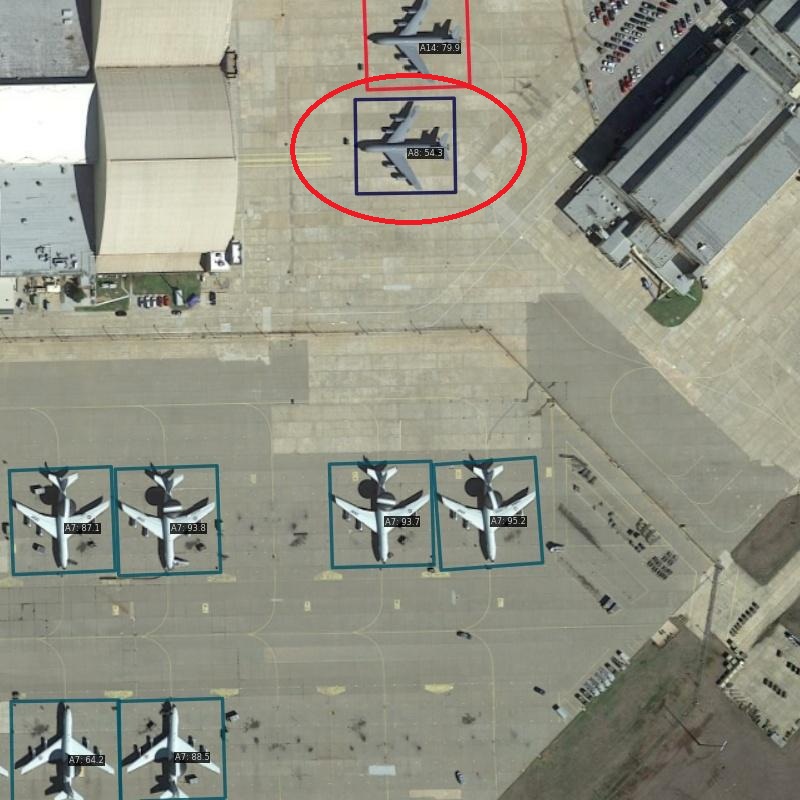}
  \end{minipage}

  \vspace{1pt}
  \centering
  \begin{minipage}[c]{.02\textwidth}
    \rotatebox{90}{(c) SPMR}
  \end{minipage}
  \begin{minipage}[c]{.157\textwidth}
    \centering
    \includegraphics[width=\textwidth]{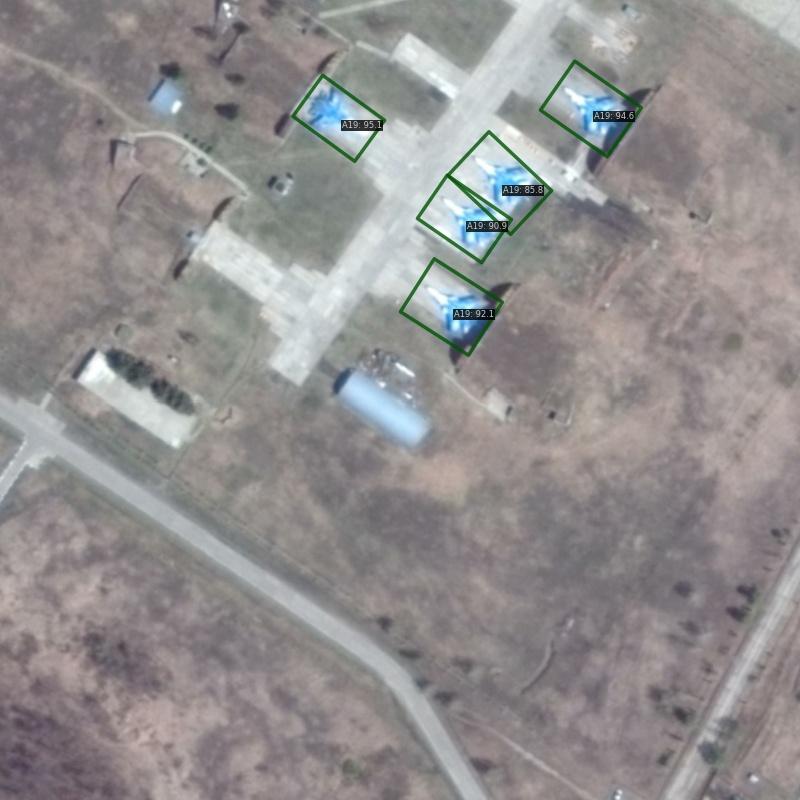}
  \end{minipage}
  \begin{minipage}[c]{.157\textwidth}
    \centering
    \includegraphics[width=\textwidth]{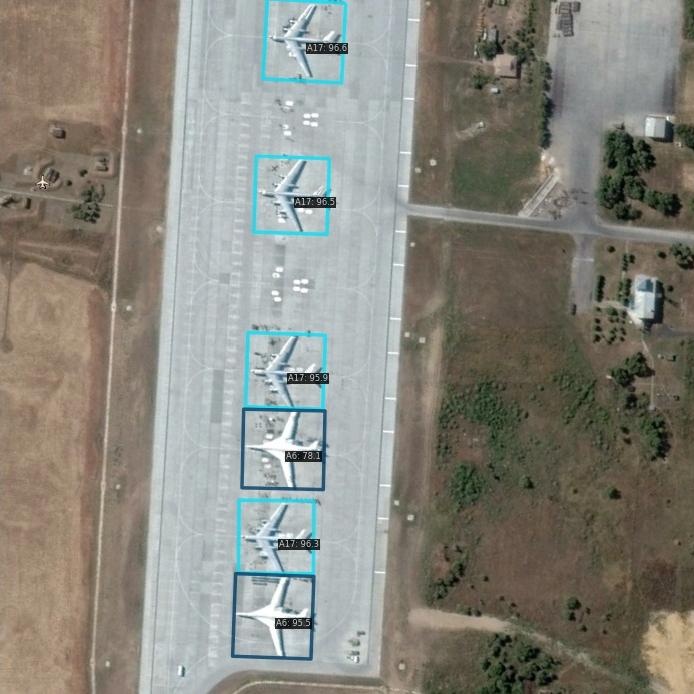}
  \end{minipage}
  \begin{minipage}[c]{.157\textwidth}
    \centering
    \includegraphics[width=\textwidth]{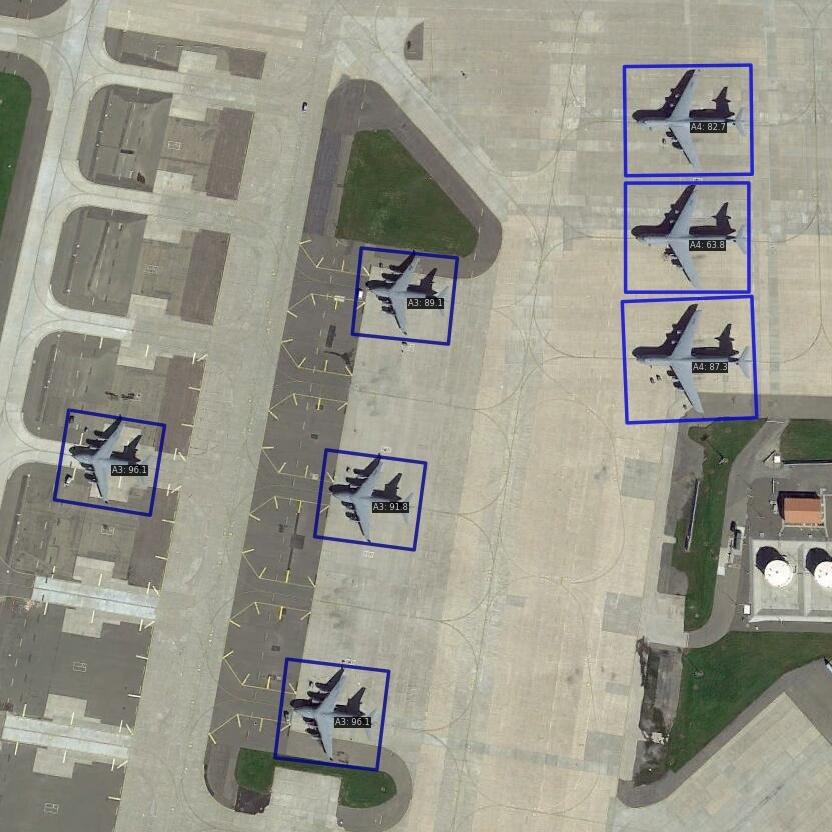}
  \end{minipage}
  \begin{minipage}[c]{.157\textwidth}
    \centering
    \includegraphics[width=\textwidth]{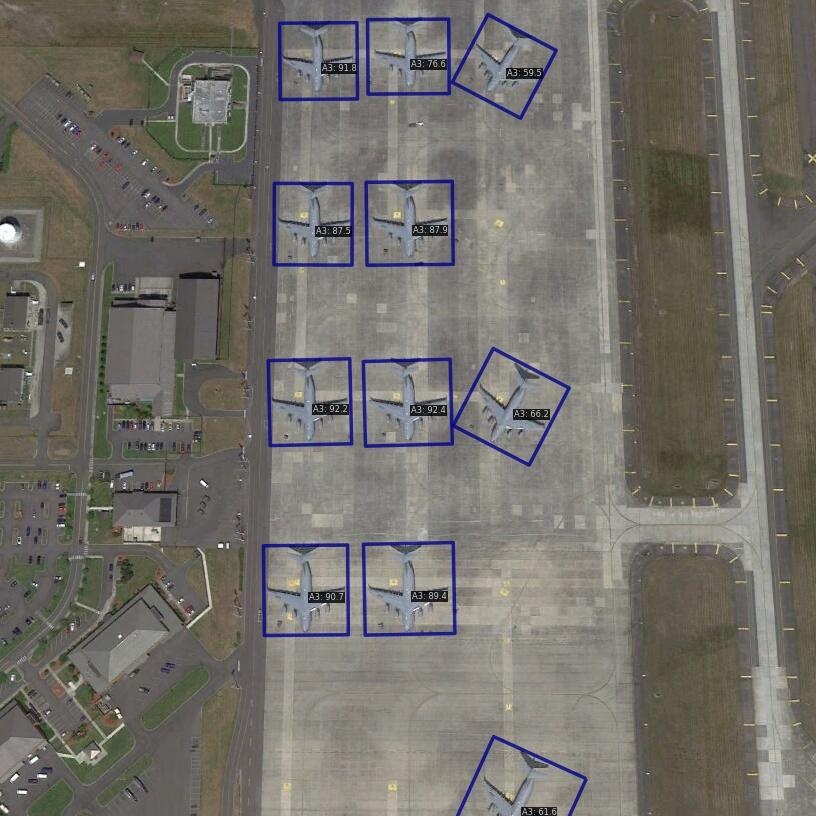}
  \end{minipage}
  \begin{minipage}[c]{.157\textwidth}
    \centering
    \includegraphics[width=\textwidth]{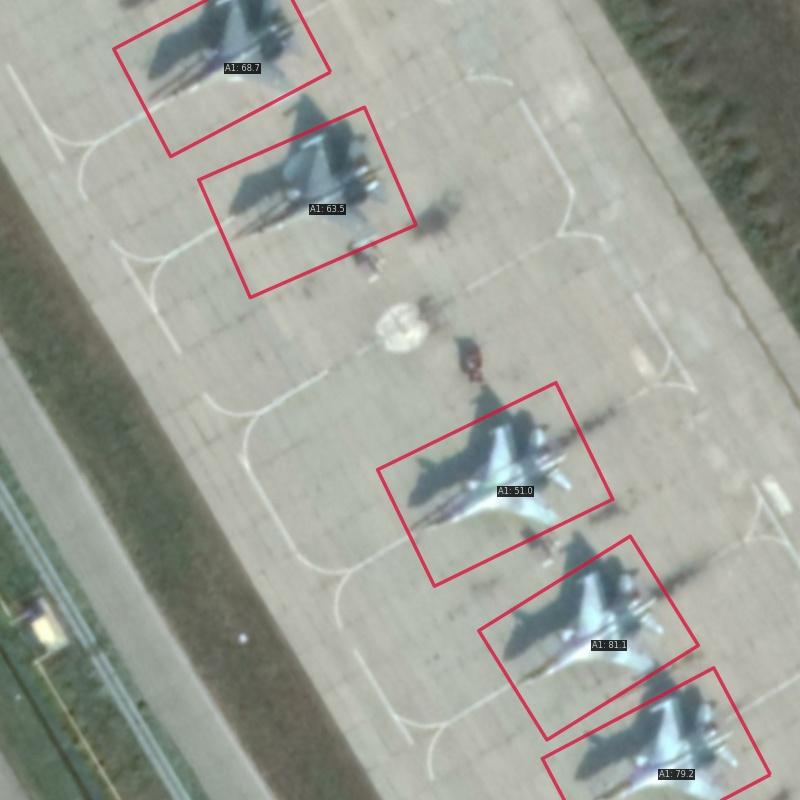}
  \end{minipage}
  \begin{minipage}[c]{.157\textwidth}
    \centering
    \includegraphics[width=\textwidth]{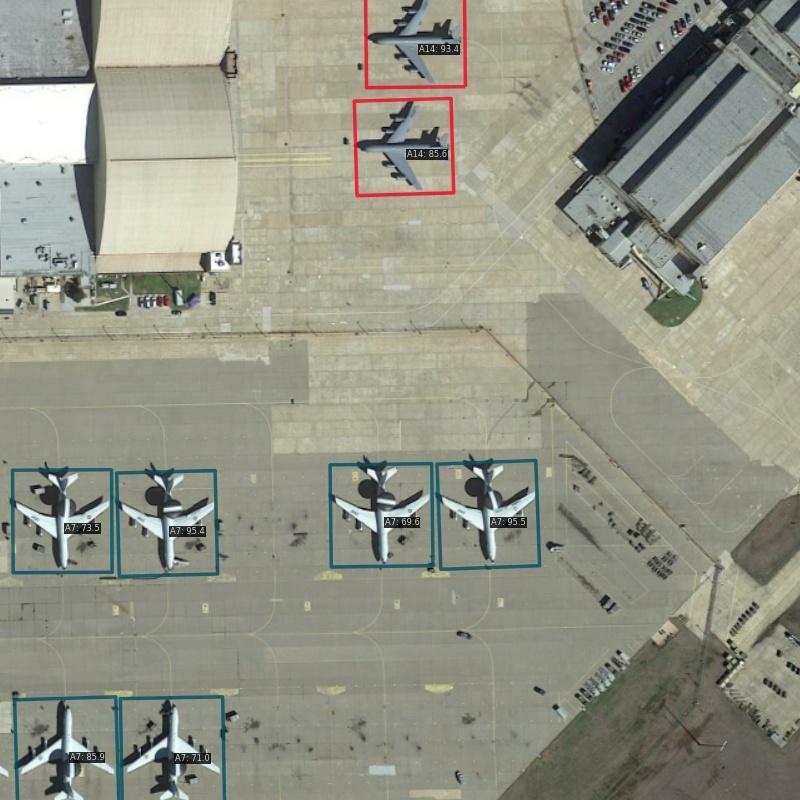}
  \end{minipage}

  \caption{Qualitative comparisons on the MAR20 test set. Ground-truth annotations are shown in the top row for reference. The second and third rows show the results of DINO-O (baseline) and the proposed SPMR, respectively. Bounding boxes in different colors indicate different fine-grained classes. The confidence threshold is set to $0.3$ for filtering predicted boxes.}
  \label{fig:qualitative_result}
\end{figure*}

\autoref{fig:qualitative_result} compares detection results on the MAR20 test set. Compared with DINO-O, SPMR reduces missed detections and misclassifications caused by semantic confusion among similar aircraft, consistent with the motivation of separating set-prediction proposal generation from proposal-level fine-grained recognition.

\subsubsection{Quantitative Results on ShipRSImageNet}
\begin{table}[!t]
  \centering
  \caption{Comparison results on ShipRSImageNet}
  \begin{tblr}{
    colsep = 5.7pt,
    column{2-5} = {c},
    cell{2}{1} = {c=4}{l},
    cell{7}{1} = {c=4}{l},
    cell{13}{1} = {c=4}{l},
    hline{3} = {-}{},
    hline{7,8} = {-}{},
    hline{13,14} = {-}{},
      }
    \toprule
    Method                          & Venue & $\mathrm{AP_{50}}$ & $\mathrm{AP_{75}}$ & $\mathrm{AP_{50:95}}$ \\
    \midrule
    \textit{One-stage methods:}                                                                                   \\
    RetinaNet-O~\cite{retinanet}    & ICCV2017  & 23.79              & 15.08              & 14.36                 \\
    \text{FCOS-O}~\cite{fcos}       & ICCV2019  & 55.76              & 46.21              & 38.24                 \\
    R$^3$Det~\cite{r3det}           & AAAI2021  & 32.82              & 20.90              & 19.51                 \\
    S$^2$ANet~\cite{s2anet}         & TGRS2022  & 52.29              & 38.74              & 32.61                 \\
    \textit{Two-stage methods:}                                                                                   \\
    Faster R-CNN-O~\cite{faster}    & NIPS2015  & 51.19              & 22.44              & 25.68                 \\
    RoI Tran~\cite{roitrans}        & CVPR2019  & 68.37              & 60.96              & 51.07                 \\
    Oriented R-CNN~\cite{orcnn}     & ICCV2021  & 69.59              & 59.48              & 49.31                 \\
    PCLDet~\cite{pcldet}            & TGRS2023  & 72.14              & 62.38              & 50.28                 \\
    PETDet~\cite{petdet}            & TGRS2024  & 74.83              & 68.99              & 56.18                 \\
    \textit{NMS-free methods:}                                                                                    \\
    Deformable-DETR-O~\cite{deform} & ICLR2021  & 69.18              & 56.92              & 48.04                 \\
    DINO-O~\cite{dino}              & ICLR2023  & 70.32              & 64.30              & 53.47                 \\
    ARS-DETR~\cite{arsdetr}         & TGRS2024  & 63.45              & 51.59              & 43.98                 \\
    OrientedFormer~\cite{oformer}   & TGRS2024  & 71.53              & 64.47              & 52.81                 \\
    RHINO~\cite{rhino}              & WACV2025  & 65.60              & 58.51              & 49.56                 \\
    \textbf{SPMR (Ours)}            & ---       & \textbf{75.40}     & \textbf{71.41}     & \textbf{60.53}        \\
    \bottomrule
  \end{tblr}
  \label{tab:shiprs}
\end{table}

The results in \autoref{tab:shiprs} further validate the generalization of SPMR on the fine-grained ship benchmark. SPMR achieves a +5.08 AP improvement over DINO-O~\cite{dino} and surpasses the strongest compared two-stage method, PETDet~\cite{petdet}, by 0.57 AP, while remaining NMS-free. These results indicate that the role-separated design transfers beyond the aircraft domain.

\subsubsection{Quantitative Results on FAIR1M}

\begin{table}[!tbp]
  \centering
  \caption{Comparison results on the FAIR1M-v1.0 and FAIR1M-v2.0 datasets. $*$ indicates results reported in the original papers. \textnormal{\textdagger} denotes multi-scale training and testing}
  \begin{tblr}{
    column{2-4} = {c},
    cell{2}{1} = {c=4}{l},
    cell{10}{1} = {c=4}{l},
    cell{17}{1} = {c=4}{l},
    hline{3} = {-}{},
    hline{10,11} = {-}{},
    hline{17,18} = {-}{},
      }
    \toprule
    Method                          & Venue & v1.0 $\mathrm{AP_{50}}$ & v2.0 $\mathrm{AP_{50}}$ \\
    \midrule
    \textit{One-stage methods:}                                                                     \\
    RetinaNet-O~\cite{retinanet}    & ICCV2017  & 25.39                   & 29.61                   \\
    \text{FCOS-O}~\cite{fcos}       & ICCV2019  & 31.68                   & 40.26                   \\
    R$^3$Det~\cite{r3det}           & AAAI2021  & 28.16                   & 34.67                   \\
    S$^2$ANet~\cite{s2anet}         & TGRS2022  & 33.80                   & 40.07                   \\
    RIDet*~\cite{ridet}             & GRSL2022  & 31.58                   & ---                     \\
    CFC-Net*~\cite{cfc}             & TGRS2022  & 34.31                   & ---                     \\
    TIOE-Det*~\cite{tioe}           & JPRS2023  & 35.16                   & ---                     \\
    \textit{Two-stage methods:}                                                                     \\
    RoI Tran~\cite{roitrans}        & CVPR2019  & 38.98                   & 44.80                   \\
    Oriented R-CNN~\cite{orcnn}     & ICCV2021  & 38.58                   & 43.82                   \\
    PCLDet~\cite{pcldet}            & TGRS2023  & 40.03                   & 44.45                   \\
    SFRNet*~\cite{sfrnet}           & TGRS2023  & 40.74                   & 45.68                   \\
    PETDet~\cite{petdet}            & TGRS2024  & 42.43                   & 48.61                   \\
    DRNet~\cite{drnet}              & TCSVT2025 & 40.04                   & 45.24                   \\
    \textit{NMS-free methods:}                                                                      \\
    Deformable-DETR-O~\cite{deform} & ICLR2021  & 37.72                   & 42.97                   \\
    DINO-O~\cite{dino}              & ICLR2023  & 39.79                   & 45.83                   \\
    ARS-DETR~\cite{arsdetr}         & TGRS2024  & 36.70                   & 42.45                   \\
    OrientedFormer~\cite{oformer}   & TGRS2024  & 38.95                   & 44.92                   \\
    RHINO~\cite{rhino}              & WACV2025  & 39.28                   & 45.13                   \\
    QEDetr*\textnormal{\textdagger}~\cite{qedetr}          & RS2025    & ---                     & 48.59                   \\
    BHCL*~\cite{bhcl}                                      & CVPR2026 & 41.66                    & 47.53                    \\
    \textbf{SPMR (Ours)}            & ---       & 42.04                   & 47.85                   \\
    \textbf{SPMR\textnormal{\textdagger} (Ours)}           & ---       & \textbf{46.34}          & \textbf{52.12}          \\
    \bottomrule
  \end{tblr}
  \label{tab:fair}
\end{table}

For the FAIR1M dataset, we conduct experiments on both FAIR1M-v1.0 and FAIR1M-v2.0, with the quantitative results presented in \autoref{tab:fair}. Under single-scale training and testing, SPMR achieves 42.04 AP on FAIR1M-v1.0 and 47.85 AP on FAIR1M-v2.0. It achieves the highest AP among the query-based methods evaluated under the same setting in \autoref{tab:fair} and remains competitive with strong two-stage FGOD methods without requiring NMS at inference.

With multi-scale training and testing, SPMR achieves 46.34 AP on FAIR1M-v1.0 and 52.12 AP on FAIR1M-v2.0, showing its applicability to large-scale scenarios.

\subsection{Ablation Study}
We conduct a set of ablation experiments to analyze the main design choices in SPMR. First, we progressively add the core components to quantify the contribution of the set-prediction and multi-positive recognition design. Subsequently, we analyze the effects and parameter settings of each component. Unless otherwise specified, all ablation studies are conducted on the MAR20 dataset using a ResNet50~\cite{resnet} backbone.

\subsubsection{Effectiveness of New Components}
\begin{table}[!t]
  \centering
  \caption{Effectiveness of each component in SPMR}
  \begin{tblr}{
    cells = {c},
    column{1} = {l},
    row{5,9} = {gray9},
      }
    \toprule
    Dataset & SOPN       & QARH       & DPAS       & $\mathrm{AP_{50}}$ & $\mathrm{AP_{75}}$ & $\mathrm{AP_{50:95}}$ \\
    \midrule
    \SetCell[r=4]{c} MAR20
            & \quad      & \quad      & \quad      & 75.36              & 62.11              & 50.65                 \\
            & \checkmark & \quad      & \quad      & 83.55              & 73.88              & 58.75                 \\
            & \checkmark & \checkmark & \quad      & 85.67              & 77.00              & 61.10                 \\
            & \checkmark & \checkmark & \checkmark & \textbf{85.82}     & \textbf{77.54}     & \textbf{61.44}        \\
    \midrule
    \SetCell[r=4]{c} ShipRS
            & \quad      & \quad      & \quad      & 70.32              & 64.30              & 53.47                 \\
            & \checkmark & \quad      & \quad      & 74.92              & 70.42              & 59.32                 \\
            & \checkmark & \checkmark & \quad      & 75.06              & 71.07              & 60.18                 \\
            & \checkmark & \checkmark & \checkmark & \textbf{75.40}     & \textbf{71.41}     & \textbf{60.53}        \\
    \bottomrule
  \end{tblr}
  \label{tab:component}
\end{table}

\autoref{tab:component} reports DINO-O as a query-based reference, followed by SPMR variants with SOPN, QARH, and DPAS added in sequence. The SOPN-only variant uses the class-agnostic proposal generator and a standard RoI head, with predictions ranked by DSCS. Adding QARH improves all three metrics on both datasets, while DPAS provides a further improvement without changing inference.
\subsubsection{Analysis of the Structure in SOPN}
\begin{table}[!t]
  \centering
  \caption{Ablation studies on different SOPN structures. FLOPs (G) and Params (M) denote computational cost in GFLOPs and parameters in millions, respectively}
  \begin{tblr}{
      cells = {c},
      hline{4}={-}{},
      row{7} = {gray9},
    }
    \toprule
    Models             & $\mathrm{AP_{50}}$ & $\mathrm{AP_{75}}$ & $\mathrm{AP_{50:95}}$ & F(G)   & P(M)  \\
    \midrule
    ORCNN~\cite{orcnn} & 82.71              & 72.15              & 58.25                 & 148.24 & 41.37 \\
    DINO-O~\cite{dino} & 75.36              & 62.11              & 50.65                 & 178.61 & 47.58 \\
    DINO-O as RPN      & 79.56              & 70.47              & 56.13                 & 173.49 & 61.31 \\
    +Decoupled Feat    & 80.64              & 72.17              & 57.41                 & 173.49 & 61.31 \\
    +2 Encoder Layers  & 81.42              & 73.20              & 58.35                 & 105.40 & 56.18 \\
    +FPN               & \textbf{83.55}     & \textbf{73.88}     & \textbf{58.75}        & 139.49 & 58.61 \\
    \bottomrule
  \end{tblr}
  \label{tab:ablation_SOPN}
\end{table}

SOPN uses an FPN to provide decoupled multi-scale features to the fine-grained stage, which reduces competition between proposal selection and later recognition. The encoder layers are reduced from six to two to keep the proposal generator lightweight and preserve the efficiency advantage of set prediction. \autoref{tab:ablation_SOPN} evaluates these design choices against Oriented R-CNN and DINO-O baselines.

In the DINO-O as RPN variant, proposals from DINO-O are passed directly to the refinement stage used in Oriented R-CNN, so the encoder features are reused for both proposal generation and refinement. This variant uses the six-layer Transformer encoder of DINO-O together with class-agnostic one-to-one proposal generation and a standard RoI head. Feature decoupling improves all three AP metrics. Reducing the encoder depth to two layers further improves AP while sharply lowering FLOPs and parameters, suggesting that a lighter encoder is sufficient once the stage interface is decoupled. Reintroducing the FPN then restores the lost multi-scale context and achieves the best overall performance.

\subsubsection{Effectiveness of QARH}
To validate the effects of Multi-Positive Recognition (MPR) and Query-Aware Decoupled Refinement (QADR) in QARH, we conduct an ablation study on top of the model equipped with SOPN. As shown in \autoref{tab:qarh}, replacing sampled RoI training with MPR while keeping the RoI-head architecture unchanged increases AP from 83.55 to 85.04. This improvement suggests that retaining all assigned proposals with quality-aware loss weighting provides more reliable fine-grained supervision. QADR alone improves AP from 83.55 to 84.72, demonstrating an independent gain over the baseline. Combining MPR with QADR achieves the highest AP, indicating that the two designs are complementary.

\begin{table}[!t]
  \centering
  \caption{Effectiveness of different designs in QARH.}
  \begin{tblr}{
      cells = {c},
      row{5} = {gray9},
    }
    \toprule
    MPR        & QADR       & $\mathrm{AP_{50}}$ & $\mathrm{AP_{75}}$ & $\mathrm{AP_{50:95}}$ \\
    \midrule
    \quad      & \quad      & 83.55              & 73.88              & 58.75                 \\
    \checkmark & \quad      & 85.04              & 76.31              & 60.42                 \\
    \quad      & \checkmark & 84.72              & 76.14              & 60.40                 \\
    \checkmark & \checkmark & \textbf{85.67}     & \textbf{77.00}     & \textbf{61.10}        \\
    \bottomrule
  \end{tblr}
  \label{tab:qarh}
\end{table}

We further compare three FFN structures in QADR. The Shared FFN applies one two-layer FFN to the fused query and RoI representation. Its output is shared by the classification and regression branches. The Decoupled FFN uses two independent two-layer FFNs to produce separate features for the two branches. All variants are trained with MPR and without DPAS, with the hidden dimension fixed at 1024. As shown in \autoref{tab:ffn_structure}, the Shared FFN uses fewer parameters than the Decoupled FFN and performs better on all three metrics. This result indicates that classification and regression still benefit from common features. The Decoupled Gated FFN keeps a shared content projection and applies separate gates for the two tasks. It gives the best results, reaching 85.67 $\mathrm{AP}_{50}$ and 61.10 $\mathrm{AP}_{50:95}$.

\begin{table}[!t]
  \centering
  \caption{Comparison of FFN structures in QADR. Params denotes the number of FFN parameters in millions.}
  \begin{tblr}{
      cells = {c},
      column{1} = {l},
      colsep = 3.5pt,
      row{4} = {gray9},
    }
    \toprule
    Structure             & Params (M) & $\mathrm{AP_{50}}$ & $\mathrm{AP_{75}}$ & $\mathrm{AP_{50:95}}$ \\
    \midrule
    Shared FFN            & 2.36       & 84.69              & 76.01              & 60.26                 \\
    Decoupled FFN         & 4.72       & 84.25              & 75.00              & 59.90                 \\
    Decoupled Gated FFN   & 6.03       & \textbf{85.67}     & \textbf{77.00}     & \textbf{61.10}        \\
    \bottomrule
  \end{tblr}
  \label{tab:ffn_structure}
\end{table}

To verify that QARH receives multi-positive supervision from the sparse SOPN proposal set, we count positive assignments among the $300$ main proposals on the MAR20 and ShipRSImageNet training sets. Here, coverage is the percentage of ground-truth instances assigned at least one positive proposal. On MAR20, the mean and median numbers of positive proposals per instance are $15.8$ and $13$, and the coverage is $97.7\%$. The corresponding values on ShipRSImageNet are $13.9$, $9$, and $92.9\%$. These statistics show that the sparse proposal set provides multiple recognition positives for typical objects while covering most ground-truth instances in both datasets. Together with the MPR improvements in \autoref{tab:qarh}, they support multi-positive supervision for QARH recognition.

\subsubsection{Impact of Query Detachment in QARH}
\begin{table}[!t]
  \centering
  \caption{Effect of query detachment in QARH. All results are reported in the QARH setting without DPAS}
  \begin{tblr}{
    cells = {c},
    cell{2}{1} = {r=2}{},
    cell{4}{1} = {r=2}{},
    hline{4} = {-}{},
    row{3,5} = {gray9},
      }
    \toprule
    Dataset        & Query Detachment & $\mathrm{AP_{50}}$ & $\mathrm{AP_{75}}$ & $\mathrm{AP_{50:95}}$ \\
    \midrule
    MAR20          & w/o detach       & 85.34              & 76.44              & 60.65                 \\
                   & w/ detach        & \textbf{85.67}     & \textbf{77.00}     & \textbf{61.10}        \\
    ShipRSImageNet & w/o detach       & 73.31              & 69.09              & 58.93                 \\
                   & w/ detach        & \textbf{75.06}     & \textbf{71.07}     & \textbf{60.18}        \\
    \bottomrule
  \end{tblr}
  \label{tab:query_detach}
\end{table}

\autoref{tab:query_detach} studies the effect of detaching SOPN queries before QARH. Query detachment consistently improves performance on both datasets, supporting the need to prevent proposal-generation queries from being directly optimized by fine-grained recognition losses.

We further investigate the robustness of the exponential scaling hyperparameter $\kappa$ in the MPR loss. Table~\ref{tab:mpr_hyper} summarizes the results under different values of $\kappa$, which controls how strongly high-quality positive proposals are emphasized according to their objectness and localization quality. Increasing $\kappa$ from $1.0$ to $2.0$ improves performance, while a further increase to $2.5$ leads to slight saturation. Overall, $\kappa=2.0$ provides the best trade-off between quality-aware supervision and stable optimization, and the results remain fairly stable across the tested range.
\begin{table}[!t]
  \centering
  \caption{Detection performance with different values of $\kappa$ in the MPR loss}
  \begin{tblr}{
      cells = {c},
      row{4} = {gray9},
    }
    \toprule
    $\kappa$ & $\mathrm{AP_{50}}$ & $\mathrm{AP_{75}}$ & $\mathrm{AP_{50:95}}$ \\
    \midrule
    $1.0$    & 84.43              & 75.76              & 60.17                 \\
    $1.5$    & 85.00              & 76.45              & 60.46                 \\
    $2.0$    & \textbf{85.67}     & \textbf{77.00}     & \textbf{61.10}        \\
    $2.5$    & 85.55              & 76.93              & 60.92                 \\
    \bottomrule
  \end{tblr}
  \label{tab:mpr_hyper}
\end{table}

\subsubsection{Impact of Consensus Weighting in DSCS}
\begin{table}[!t]
  \centering
  \caption{Detection performance with different values of $\alpha$ in DSCS}
  \begin{tblr}{
      cells = {c},
      row{4} = {gray9},
    }
    \toprule
    $\alpha$ & $\mathrm{AP_{50}}$ & $\mathrm{AP_{75}}$ & $\mathrm{AP_{50:95}}$ \\
    \midrule
    $0.5$    & 84.61              & 76.25              & 60.56                 \\
    $0.6$    & 85.40              & 77.13              & 61.19                 \\
    $0.7$    & 85.82              & \textbf{77.54}     & \textbf{61.44}        \\
    $0.8$    & \textbf{85.88}     & 77.39              & 61.35                 \\
    $0.9$    & 85.85              & 77.27              & 61.24                 \\
    \bottomrule
  \end{tblr}
  \label{tab:dscs_alpha}
\end{table}

\autoref{tab:dscs_alpha} presents a sensitivity analysis for the hyperparameter $\alpha$ in DSCS. Here, $\alpha$ balances the contributions of the SOPN objectness score and QARH fine-grained classification score before ranking. The results show that the sparse objectness prior is important for final ranking, with stable performance in the range $\alpha \in [0.7, 0.9]$. We adopt $\alpha=0.7$ as a fixed default for all datasets, rather than tuning it separately for each dataset.

\subsubsection{Efficacy and Efficiency of NMS-free Inference}
\begin{table}[!t]
  \centering
  \caption{Comparison of scoring and suppression strategies}
  \begin{tblr}{
      cells = {c},
      row{4} = {gray9},
    }
    \toprule
    Scoring & NMS        & $\mathrm{AP_{50}}$ & $\mathrm{AP_{75}}$ & $\mathrm{AP_{50:95}}$ & FPS           \\
    \midrule
    QARH          & \checkmark & 84.21              & 74.92              & 59.47                 & 9.4           \\
    DSCS          & \checkmark & \textbf{85.89}     & 77.09              & 61.19                 & 9.3           \\
    DSCS          &            & 85.82              & \textbf{77.54}     & \textbf{61.44}        & \textbf{15.9} \\
    \bottomrule
  \end{tblr}
  \label{tab:dscs_nms}
\end{table}

\autoref{tab:dscs_nms} compares scoring and suppression strategies. Here, \emph{QARH} denotes direct use of the QARH fine-grained score, whereas \emph{DSCS} denotes consensus scoring with SOPN objectness and QARH class scores. Rotated NMS is applied to the first two variants, and the last variant uses DSCS directly for NMS-free inference. The NMS threshold is fixed to $0.1$ for all NMS-based variants. Under the same NMS setting, DSCS improves all three AP metrics over QARH score ranking, indicating that SOPN objectness provides complementary information for ranking the refined proposals. Adding rotated NMS on top of DSCS gives a marginally higher $\mathrm{AP_{50}}$ but lower $\mathrm{AP_{75}}$ and $\mathrm{AP_{50:95}}$. Direct NMS-free inference with DSCS achieves the highest inference speed. SPMR therefore uses DSCS as the final ranking rule without pairwise suppression.

\section{Conclusion} \label{sec:conclusion}
In this work, we revisited NMS-free fine-grained object detection in remote sensing from the perspective of role separation between sparse proposal generation and fine-grained recognition. SPMR retains one-to-one set prediction for class-agnostic oriented proposal generation and duplicate suppression, while introducing proposal-level multi-positive recognition to provide richer supervision for fine-grained discrimination. Experiments on MAR20, ShipRSImageNet, and FAIR1M show that SPMR achieves competitive performance among FGOD detectors and consistent gains over representative NMS-free oriented detectors. These results suggest that NMS-free FGOD benefits from separating duplicate-suppressed proposal generation from multi-positive fine-grained recognition and calibrating their scores for final ranking. The current design also has several practical limitations. The refinement stage adds computation, and recognition performance depends on the recall and localization quality of SOPN proposals. DSCS uses the same fusion weight for all predictions. Future work will explore simpler role-separated designs and adaptive score fusion to improve efficiency and flexibility.

\ifCLASSOPTIONcaptionsoff
  \newpage
\fi
\printbibliography

\end{document}